\documentclass{article}

\usepackage{microtype}
\usepackage{graphicx}
\usepackage{booktabs} %

\usepackage{hyperref}

\usepackage[accepted]{icml2025}

\usepackage{amsmath}
\usepackage{amssymb}
\usepackage{mathtools}
\usepackage{amsthm}

\usepackage{subcaption}
\usepackage{MnSymbol}
\usepackage{wrapfig}
\usepackage{xcolor}
\usepackage{arydshln}
\usepackage{multirow}
\usepackage{listings}
\usepackage{adjustbox}
\usepackage{float}
\usepackage{lipsum}
\usepackage{algorithm}  
\usepackage[algo2e]{algorithm2e}

\newcommand{\ours}{MaGNeTS}

\definecolor{codegreen}{rgb}{0,0.4,0}
\definecolor{codegray}{rgb}{0.,0.,0.}
\definecolor{codepurple}{rgb}{0.45,0,0.6}
\definecolor{backcolour}{rgb}{0.96,0.95,0.92}

\lstdefinestyle{mystyle}{
    backgroundcolor=\color{backcolour},   
    commentstyle=\color{codegreen},
    keywordstyle=\color{magenta},
    numberstyle=\tiny\color{codegray},
    stringstyle=\color{codepurple},
    basicstyle=\ttfamily\footnotesize,
    breakatwhitespace=false,         
    breaklines=true,                 
    captionpos=b,                    
    keepspaces=true,                 
    numbers=left,                    
    numbersep=5pt,                  
    showspaces=false,                
    showstringspaces=false,
    showtabs=false,                  
    tabsize=2
}

\usepackage[capitalize,noabbrev]{cleveref}

\theoremstyle{plain}

\theoremstyle{definition}

\theoremstyle{remark}

\usepackage[textsize=tiny]{todonotes}

\icmltitlerunning{\ours}

\begin{document}

\twocolumn[
\icmltitle{Masked Generative Nested Transformers with Decode Time Scaling}

\icmlsetsymbol{equal}{*}
\icmlsetsymbol{google}{†}

\begin{icmlauthorlist}
\icmlauthor{Sahil Goyal}{equal,comp}
\icmlauthor{Debapriya Tula}{equal,google,sch}
\icmlauthor{Gagan Jain}{comp}
\icmlauthor{Pradeep Shenoy}{comp}
\icmlauthor{Prateek Jain}{comp}
\icmlauthor{Sujoy Paul}{equal,comp}
\end{icmlauthorlist}

\icmlaffiliation{comp}{Google DeepMind}
\icmlaffiliation{sch}{University of California, Los Angeles}

\icmlcorrespondingauthor{Sahil Goyal}{goyalsahil@google.com}
\icmlcorrespondingauthor{Sujoy Paul}{sujoyp@google.com}

\icmlkeywords{Machine Learning, ICML}

\vskip 0.3in
]

\printAffiliationsAndNotice{\icmlEqualContribution, \workatgoogle} %

\begin{abstract}
Recent advances in visual generation have made significant strides in producing content of exceptional quality. However, most methods suffer from a fundamental problem - a bottleneck of inference computational efficiency. Most of these algorithms involve multiple passes over a transformer model to generate tokens or denoise inputs. However, the model size is kept consistent throughout all iterations, which makes it computationally expensive. In this work, we aim to address this issue primarily through two key ideas - (a) not all parts of the generation process need equal compute, and we design a decode time model scaling schedule to utilize compute effectively, and (b) we can cache and reuse some of the computation. Combining these two ideas leads to using smaller models to process more tokens while large models process fewer tokens. These different-sized models do not increase the parameter size, as they share parameters. We rigorously experiment with ImageNet256$\times$256 , UCF101, and Kinetics600 to showcase the efficacy of the proposed method for image/video generation and frame prediction. Our experiments show that with almost $3\times$ less compute than baseline, our model obtains competitive performance.
\end{abstract}

\section{Introduction}

The last decade has witnessed tremendous progress in image andvideo generation, under diverse paradigms - Generative Adversarial Networks \citep{brock2018large,sauer2022styleganxlscalingstyleganlarge}, denoising processes such as diffusion models \citep{ho2020denoising, ho2022video, dhariwal2021diffusion,ldm,vqdiffusion}, image generation via vector quantized tokenization \citep{razavi2019generating,esser2021taming,ge2022longvideogenerationtimeagnostic,van2017neural}, and so on. In recent years, diffusion models and modeling visual tokens as language have been the de-facto processes used to generate high-quality images. While initially proposed with a CNN or U-Net based architectures \citep{ldm,saharia2022photorealistic}, transformer models have become the norm now for these methods \citep{peebles2023scalablediffusionmodelstransformers,yu2023magvit}.

\begin{figure*}[h]
    \centering
    \includegraphics[width=0.8\textwidth]{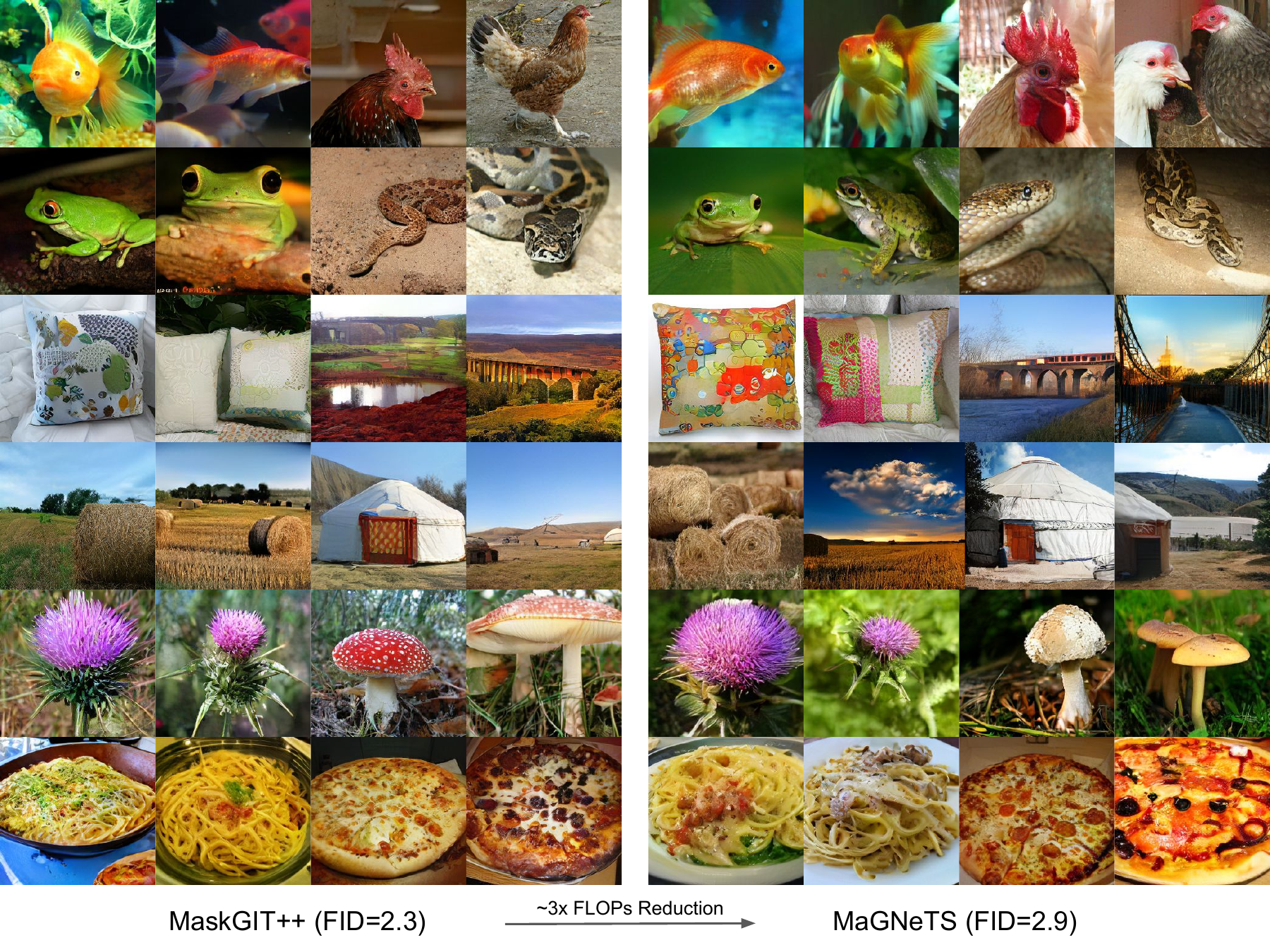}
    \caption{\small \textbf{Class-conditional image generation on ImageNet256x256.}. Comparing MaskGIT++ and \ours\ (size: L). }
    \label{fig:qual_img}
    \vspace{-1mm}
\end{figure*}

The recent advancements in visual generation can be categorized along two axes -- (a) different types of denoising processes in the continuous latent space \citep{ho2020denoising,nichol2021improved}, discrete space \citep{vqdiffusion,loudiscrete} or masking in the discrete space \cite{yu2023magvit,chang2022maskgit}, continuous space \citep{li2024autoregressive} (b) modeling tokens either auto-regressively \citep{kondratyuk2023videopoet,esser2021taming,yu2021vector} with causal attention or parallel decoding with bi-directional attention \citep{vqdiffusion,yu2023magvit,chang2022maskgit, zheng2022movqmodulatingquantizedvectors}. To achieve a high synthesis fidelity, both, denoising in diffusion models, and raster scan based auto-regressive token modeling require several iterations.

Recently, parallel decoding of discrete tokens have shown promise in generating high quality images with few iterations - MaskGIT \citep{chang2022maskgit}, MAGVIT \citep{yu2023magvit}, MUSE \citep{chang2023muse}, MaskBIT \citep{weber2024maskbitembeddingfreeimagegeneration}, TiTok \citep{yu2024imageworth32tokens}. These models are trained with Masked Language Modeling (MLM) type losses, and the generation process involves unmasking a few confident tokens every decoding iteration, starting from all masked tokens. They can even surpass diffusion models, given a good visual tokenizer \citep{yu2023language, weber2024maskbitembeddingfreeimagegeneration}.

Although MaskGIT reduces decode complexity significantly, parallel decoding still includes several redundant computations. First, the need for same capacity model for all steps needs to be investigated. Second, unlike auto-regressive models, which cache its computation in all steps, parallel decoding models perform re-computation for all tokens. We empirically find that a smaller model can generate good-quality images but its performance saturates after a point with more decoding iterations. A bigger model can perform finer refinement and generate better-quality images.

Motivated by these observations, we present \textbf{Ma}sked \textbf{G}enerate \textbf{Ne}sted \textbf{T}ransformers with Decode Time \textbf{S}caling (\ours). We design a model size curriculum over the decoding process, which efficiently utilize compute. \ours \ gradually scales the model size up to the full model size over the decoding iterations instead of using a single large model throughout. Operating on discrete tokens, we cache key-value pairs of unmasked tokens and reuse them in later iterations. A combined effect of these two techniques leads to processing more tokens with smaller and fewer tokens with larger models. The heterogenous sized models share parameters, sequentially occupying larger and larger subspaces of the parameter space, as in MatFormer \citep{kudugunta2023matformer}. We build \ours \ on top of MaskGIT.  We find that MaskGIT can be drastically improved using classifier-free guidance, specifically when trained with it. We call this MaskGIT++ and use this as the improved baseline, presenting all results on top of it. 

On ImageNet, with $\sim3\times$ less compute, \ours \ generates images of similar quality as MaskGIT++ (see \cref{fig:qual_img}). It is also comparable to state-of-the-art methods, which need orders of magnitude more compute. We also show \ours's efficacy on video datasets like UCF101 \citep{soomro2012ucf101dataset101human} and Kinetics600 \citep{carreira2018shortnotekinetics600}. To summarize, the main contributions of this work are:
\begin{itemize}
\itemsep0em 
    \item We introduce the concept of model size scheduling during the generation process to significantly reduce compute requirements.
    \item We show that like auto-regressive models, KV-caching can also be used in parallel decoding, which can effectively reuse computation when refreshed appropriately.
    \item We introduce nested modeling in image/video generation to exploit the above ideas effectively.
    \item Extensive experiments show that \ours \ offers $2.5$ - $3.7\times$ compute gains across tasks.
\end{itemize}

\section{Related Work}

\paragraph{Efficient Visual Generation.}
Image generation literature has seen significant improvements in the past years - Generative Adversarial Networks \citep{brock2018large,sauer2022styleganxlscalingstyleganlarge}, discrete token based models \citep{chang2022maskgit, yu2023magvit}, diffusion-based models \citep{kingma2023understanding, hoogeboom2023simple}, and more recently hybrid models \citep{peebles2023scalablediffusionmodelstransformers, yu2024representationalignmentgenerationtraining}, but they often guzzle computing power. Researchers tackle this bottleneck of computational costs with efficient model architectures and smarter sampling strategies.

In  diffusion model literature, there have been some work to reduce the number of sampling steps, by treating the sampling process like ordinary differential equations \cite{song2022denoisingdiffusionimplicitmodels, lu2022dpmsolverfastodesolver, liu2022pseudonumericalmethodsdiffusion}, incorporating additional training process \cite{kong2021fastsamplingdiffusionprobabilistic, nichol2021improveddenoisingdiffusionprobabilistic, salimans2022progressivedistillationfastsampling, song2023consistencymodels}, sampling step distillation \cite{salimans2022progressivedistillationfastsampling, song2023consistencymodels,berthelot2023tractdenoisingdiffusionmodels,meng2023distillationguideddiffusionmodels, feng2024relationaldiffusiondistillationefficient}, sampling and training formulation modifications \cite{esser2024scalingrectifiedflowtransformers,song2023consistencymodels}, and more. 
Recently, there has been growing interest in understanding how each step in the diffusion sampling process contributes \cite{ choi2022perceptionprioritizedtrainingdiffusion, park2023understandinglatentspacediffusion, betasampling}. %
These approaches analyze sampling steps leveraging distance metrics such as LPIPS, Fourier analysis, and spectral density analysis.
Building on these explorations researchers have designed methods based on optimal sampling steps \cite{watson2022learningfastsamplersdiffusion,betasampling}, weighted training loss \cite{choi2022perceptionprioritizedtrainingdiffusion}, and step-specific models \cite{li2023autodiffusiontrainingfreeoptimizationtime, yang2024denoisingdiffusionstepawaremodels, lee2023multiarchitecturemultiexpertdiffusionmodels}. These step-specific models use computationally expensive evolutionary search algorithms, directly optimizing the quality metric, FID. Concurrently, researchers are actively addressing the inherent architectural costs of diffusion models, particularly those associated with transformer attention mechanisms \cite{yuan2024ditfastattn,yan2024diffusion}.
 
On the other hand, certain works focus on building better tokenizers. \citet{ldm} took diffusion models from pixel to compressed latent space for efficient and scalable generation.  \cite{yu2023language,weber2024maskbitembeddingfreeimagegeneration} explore certain vector quantizers in the tokenization process to improve generation quality. \citet{tian2024visualautoregressivemodelingscalable} explore multi-scale tokenizer to improve quality while \citet{yu2024imageworth32tokens} looks at 1D tokenizers to reduce the number of compressed tokens. Instead of sampling or tokenization process optimization, we tackle an orthogonal problem of efficient compute allocation over the multi-step generation process. This makes our approach usable with a variety of tokenizers, model architectures and sampling schemes.

{\flushleft \textbf{Nested Models.}} \citet{rippel2014learning} introduced nested dropout to learn ordered representations of data that improve retrieval speed and adaptive data compression. Matryoshka Learning \citep{kusupati2022matryoshka} introduces the concept of nested structures into embedding dimensions of the transformer, making them flexible. MatFormer \citep{kudugunta2023matformer} applies
the same concept to the MLP hidden layer in each transformer block. Recent methods like \citep{cai2024matryoshka, hu2024matryoshka} explore the idea of nested models in multimodal large language models. MoNE \citep{jain2024mixturenestedexpertsadaptive} and Flextron \citep{cai2024flextron} learn to route tokens to variable sized nested models  leading to compute efficient processing. In this work, we show how different parts of a multi-step task like image generation, can be modeled by different sized nested models instead of always decoding via the full model, thus significantly reducing computations.

\begin{figure*}[t]
    \centering
    \includegraphics[width=0.8\textwidth]{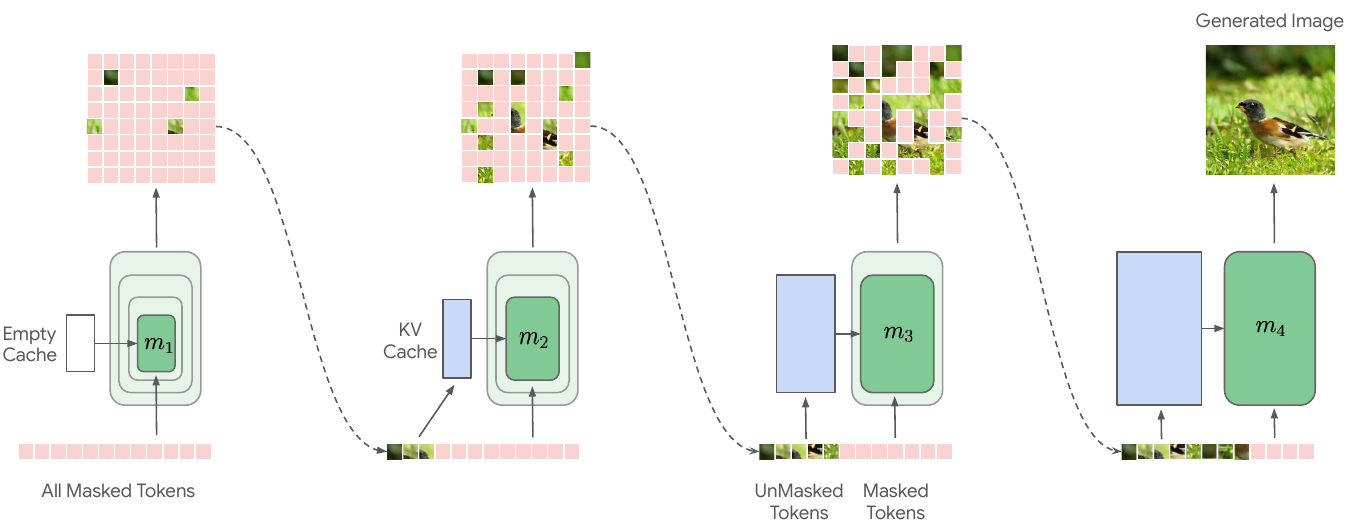}
    \caption{\small \textbf{\ours \ Decoding.} We start from the smallest nested model with an empty cache and gradually move to bigger models over the decoding iterations. We iterate using a particular sized model for a few iterations, before moving onto the next model size. As we cache the key-value pairs for the unmasked tokens, the KV cache size also increases over time. We also refresh the cache when we switch models, hence its dimension also increases over decoding iterations.}
    \label{fig:main_algo}
\end{figure*}

\begin{figure*}[h]
    \centering
    \includegraphics[trim={4.5cm 3cm 4cm 8cm},   width=0.95\textwidth]{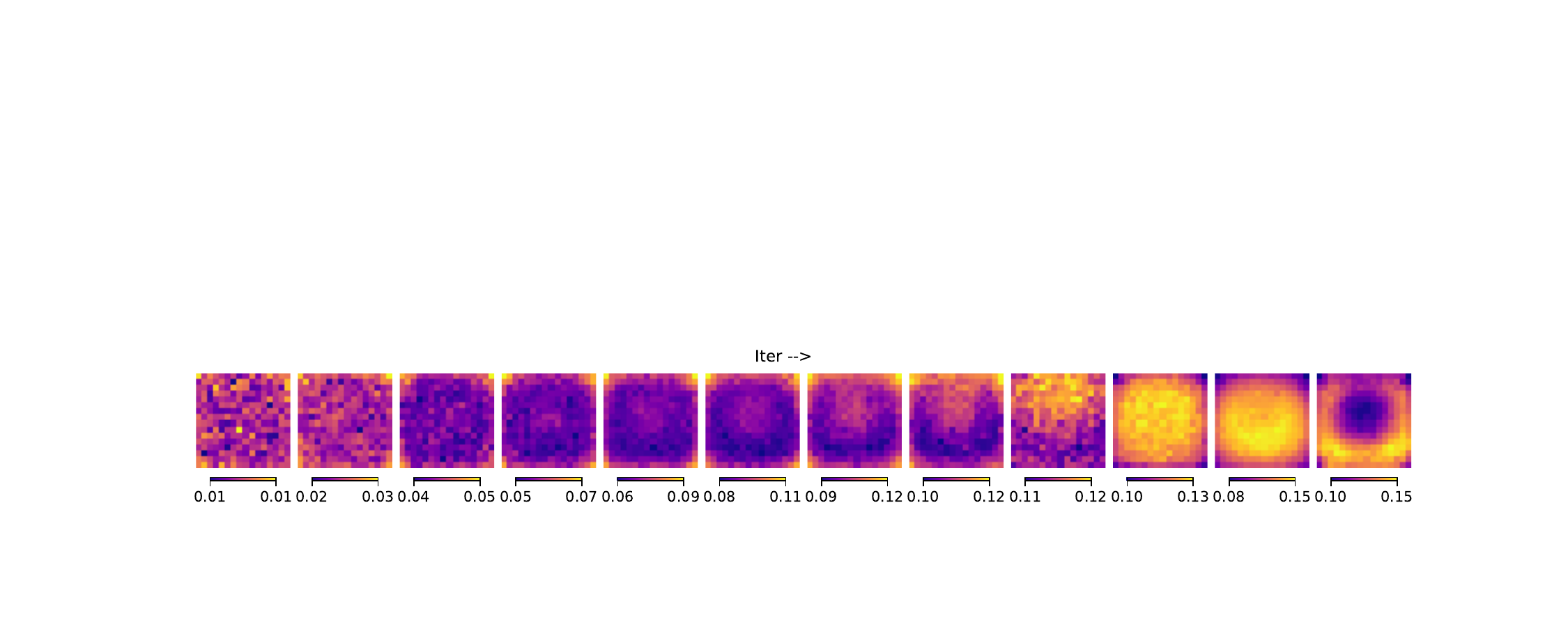}
    \caption{\small \textbf{Unmasked Token Density} visualization in each decoding iteration averaged over 50k generated samples on ImageNet. Yellow represents higher density. Each pixel represent a token from $16 \times 16$ latent token space. (See \cref{appendix:motivation} for category-wise token density).}
    \label{fig:vis_tokens}
\end{figure*}

\section{Preliminaries} \label{subsec:prelim}

{\flushleft \textbf{Parallel Decoding for Image Generation.}}
Masked Generative Image Transformer (MaskGIT) \citep{chang2022maskgit} introduces a novel approach to image generation that significantly differs from traditional autoregressive models. In autoregressive decoding, images are generated sequentially, one pixel/token at a time, following a raster scan order \citep{esser2021taming,kondratyuk2023videopoet, wang2024parallelizedautoregressivevisualgeneration, yu2024randomizedautoregressivevisualgeneration, li2024autoregressiveimagegenerationvector}. This sequential approach can be computationally inefficient, as each token is conditioned only on the previously generated tokens, leading to a bottleneck in processing time. MaskGIT generates all tokens of an image simultaneously and then iteratively refines them. This method enables significant acceleration in the decoding process. The tokens are discrete and obtained using Vector Quantized (VQ) autoencoders, learned with self-reconstruction and photo-realism losses \citep{yu2023magvit}. The iterative parallel decoding process can be represented as:
\begin{equation}
    \mathbf{X}_{k} \leftarrow \mathrm{Mask \circ Sample}({M}({\mathbf{X}}_{k-1}, c), k)
    \label{eqn:maskgit}
\end{equation}
where $\mathbf{X} \in \mathbb{Z}^N_{\geq 0}$, are the input tokens, $N$ is the number of tokens, $k \in [1, K]$ denote the iteration number, with $K$ being the total number of iterations, $\mathbf{X}_0$ is either completely masked for full generation, and partially masked for conditional generation tasks like frame prediction, $c$ is the category of image/video under generation. The $\mathrm{Sample}$ function utilizes logits predicted by the model ${M(.)}$, introduces certain randomness, and sorts them by confidence, unmasking only top-k tokens while masking the rest. We follow this process as in \cite{chang2022maskgit,yu2023magvit}. 

{\flushleft \textbf{Nested Models.}}
The core of our algorithm for inference-efficient decoding relies on variable-sized nested models for efficient parameter-sharing and hence feature space. We use MatFormer's \citep{kudugunta2023matformer} modeling approach to extract multiple nested models, from a single model, without increasing the total parameter count. Given a full transformer model $M$, MatFormer defines nested models $\{m_1, \dots, m_C\}$, such that $m_1 \subset m_2 \dots \subset m_C = M$. Each $m_i$ has fewer parameters and reduced compute. The core idea of extracting nested models is that in a transformer block, a reduced computation using a parameter subspace can be performed via a sliced matrix multiplication. Assuming a parameter matrix $\mathbf{W} \in \mathbb{R}^{d'\times d}$ and feature vector $\mathbf{x} \in \mathbb{R}^d$, then the computation $\mathbf{y}=\mathbf{W}\mathbf{x}$ can be partially obtained by computing $\mathbf{y}_{[:\frac{d'}{p}]}=\mathbf{W}_{[:\frac{d'}{p}, :]}\mathbf{x}$, if $\mathbf{y}$ is desired to be partial and $\mathbf{y}=\mathbf{W}_{[:, :\frac{d}{p}]}\mathbf{x}_{[:\frac{d}{p}]}$, if input $\mathbf{x}$ is partial. Nested models can be obtained via partial computations throughout the network.

While MatFormer \citep{kudugunta2023matformer} obtained sub-models with partial computation only in the MLP layer, we also do it in the Self-Attention layer, specifically in obtaining the $\mathbf{Q}, \mathbf{K}, \mathbf{V}$ features. These features are of dimension $n_h \times \frac{d_h}{p}$, where $n_h$ is the number of attention heads, $d_h$ is the head feature dimension, and $p$ is the model downscaling factor. We choose four downscaled models $C=4$, with $p \in \{1, 2, 4, 8\}$ in this work. After attention computation, this gives us features that are also $p$ times downscaled. Then, it is projected back to the full model dimension $d$ using partial computation, as the input features are partial. The same strategy is applied to the MLP layer. This process gives us models with close to linear reduction in parameter count and inference compute with the downscaling factor $p$.

\section{Method}
Given the preliminaries, here we introduce the core algorithm. We first discuss the idea of scheduling models of different sizes over decode iters of MaskGIT. Then, we discuss the process of caching key-value in parallel decoding, followed by how to refresh them to improve performance. We finally discuss the nested model training method. A pictorial overview of our method is presented in \cref{fig:main_algo}.

{\flushleft \textbf{Decode Time Model Schedule.}}
In iterative parallel decoding \citep{chang2022maskgit,yu2023magvit}, the same-sized model is used for all steps, starting with all tokens being masked. However, we hypothesize that certain stages of the generation process might be easier than others. For example, in the initial steps, the model only needs to capture coarse global structures, which can be achieved efficiently using smaller models. In the later steps, the model must refine finer details, which requires larger models. This hypothesis is bolstered with \cref{fig:vis_tokens}, which shows that the generation process starts unmasking tokens from the background and shifts to the middle of the image in the later iterations (more categorical examples in Appendix \cref{fig:vis_tokens_supp}). 

\begin{figure}[!h]
    \centering
    \includegraphics[scale=0.24]{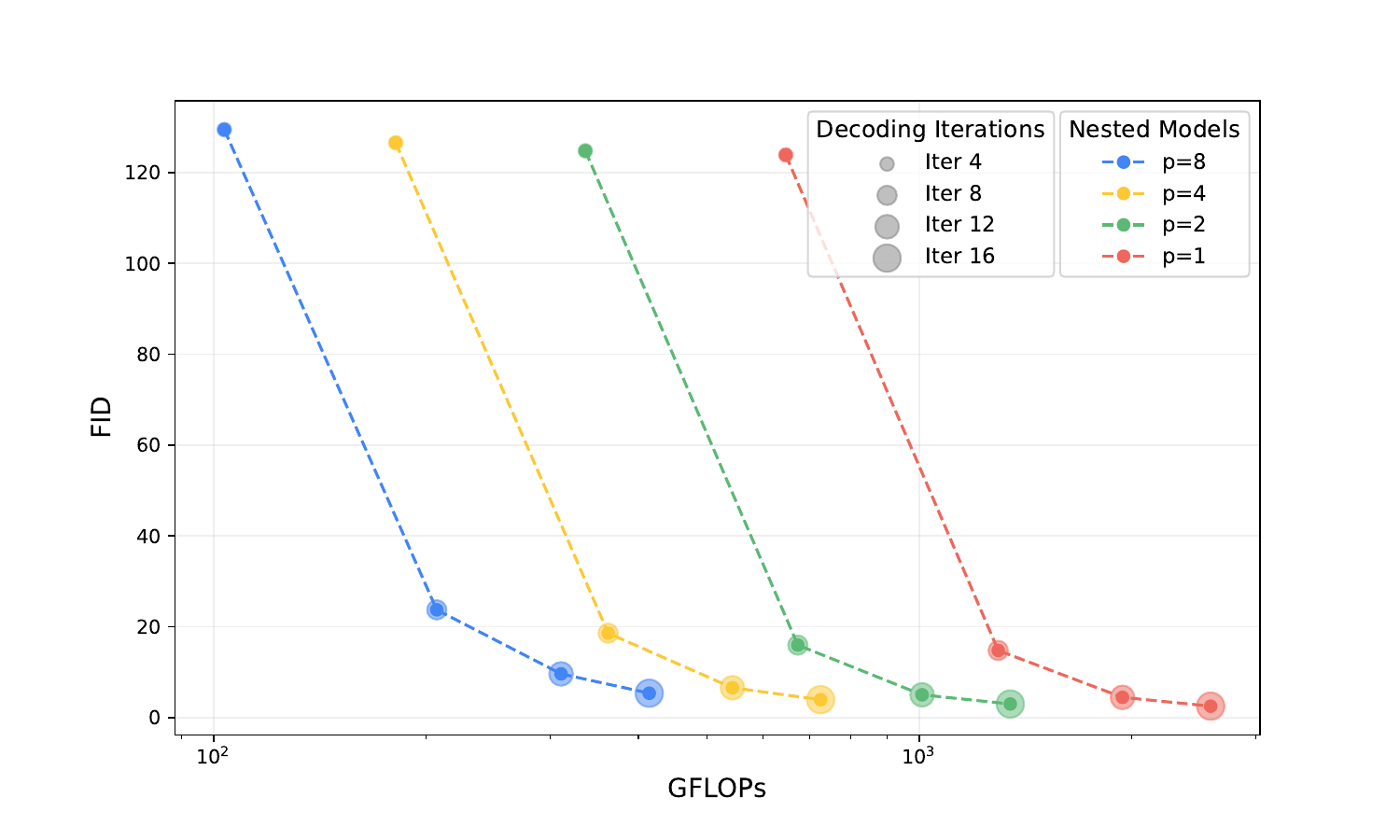}
    \caption{\small \textbf{Nested Models at different decoding iterations}. Different values of the downscaling factor $p$ correspond to the nested models. The diameter of the blobs indicates \#iterations.}
    \label{fig:fid_over_iter}
\end{figure}

Our hypothesis is further motivated by \cref{fig:fid_over_iter}, which presents the generation quality (FID) over iterations of parallel decoding for different-sized models. The smallest model reaches a reasonably good FID score with very low FLOPs compared to the biggest model. However, it saturates after a point, and the larger models surpasses the smaller ones in performance, demonstrating their ability to capture finer details and generate higher-quality images when provided with sufficient compute. This trend suggests that dynamically scaling the model size during decoding can exploit the varying task difficulty and achieve compute efficiency.

We use nested models to extract multiple models rather than using models with disjoint parameters. Nested models do not increase the parameter count and it also helps in better alignment of hypothesis when we shift model size over decode steps. The decode time model schedules can be generalized and represented as making the model choice in \cref{eqn:maskgit} dependent on the iteration index as follows: 
\vspace{-5mm}

{\small
\begin{align}
    \mathbf{X}_{k} &\leftarrow \mathrm{Mask \circ Sample}(\mathcal{M}_{k}({\mathbf{X}}_{k-1}, c), k) \nonumber \\
    \mathcal{M} = \{&(m_{p_1})^{k_1}, (m_{p_2})^{k_2}, \dots, (m_{p_n})^{k_n}\}, \ \  \text{s.t.} \sum_{i}^{n} k_i = K
    \label{eqn:schedule}
\end{align}}
where $p_1, p_2, \dots, p_n$ denoting the downscaling factors of the corresponding nested models, and $(m)^k$ denotes that model $m$ will be executed for $k$ iterations. $K$ represents the total number of iterations. We can think of different model schedules -  downscaling (starting with the full model and then gradually moving to the smallest model), upscaling, intermittently switching among a few models, and so on. We can also modify the integers $k_i$ to choose the number of times we stick to a model before switching. However, as intuitively discussed before, we empirically validate that gradually upscaling the model size gets the best trade-off between the compute and generation quality. 

{\flushleft \textbf{Cached Parallel Decoding.}}
Inspired by caching key-value pairs in auto-regressive models, we explore caching in parallel decoding, which retains relevant computations and enhances efficiency.
In auto-regressive models, caching progressively happens in one fixed direction. However, in parallel decoding, caching must depend on which tokens are unmasked over the iterations. 

Concretely, starting from an empty cached set, we keep adding keys and values to the set for the tokens that are unmasked after the $\mathrm{Mask} \circ \mathrm{Sample}$ steps (see \cref{subsec:prelim}). We do not update the predicted token indices for these unmasked tokens. Hence, the cached key and values for the unmasked tokens are the only features the other masked tokens need; hence, we do not need any further computation. In every decoding iteration, we can categorize tokens into three main categories: unmasked tokens (for which we have cached KV), tokens that can be unmasked during the current iteration, and the rest of the tokens. Note that the KV cache for the second category tokens cannot be used in the next iteration but only in the iteration after that once we know their token indices after the forward pass. We cache them in the next iteration for use in the immediately next iteration.

Caching is even more useful for decode time model schedules. For a schedule that progressively scales up the model size as decoding progresses, smaller models process more tokens, while the larger models process fewer tokens, leading to an efficient yet good quality image generation process.

\begin{algorithm}
\tiny
\caption{\small \ours \ Decoding Algorithm}
\label{alg:our_algo}
\KwIn{ 
    \( \mathbf{X}^0 \) (Initial Tokens), 
    \( K \) (\#steps), 
    \( N \) (\#tokens),
    \(\mathcal{M} \) (Nested Model Schedule),
     \( c \) (class),
    \\
}

Initialize: \( k \gets 0 \); \( {cache} \gets \{\} \); 

Note:  \(\mathbf{X}^0\) is a list of token ids  (Mask token id = \(-1\))

\While{step \(k < K\)}{
    \If{ \(k > 0\) and \(\mathcal{M}_k \neq \mathcal{M}_{k-1} \)} {Clear \( {cache} \)}
    
    Get uncached tokens: \( \mathbf{X}_{uc}^{k} \gets \{x_i \;|\; x_i \in \mathbf{X}^{k};\ i \notin cache\} \)

    Compute prediction probabilities and key-values: \( p^{k}, (kv) \gets \mathcal{M}_{k}(\mathbf{X}_{uc}^{k}; cache)\) %

    Sample tokens using current predictions $p^k$, without modifying previous predictions,
    \[\mathbf{X}^{k+1} \gets \text{MaskGIT-Sample}(p^k) %
    \]

    New indices to cache: \(\mathcal{C} \gets \{i \;|\ i \notin cache,\ \mathbf{X}^{k+1}_i\neq -1\}\)
    
    Update the kv cache: \( cache \gets cache \;\bigcup \;\{i:(kv)_i \;|\; i \in \mathcal{C}\}\)

    \( k \gets k + 1 \)
}
\KwRet $\mathbf{X}^K$
\end{algorithm}

{\flushleft \textbf{Intermittent Cache Refresh.}}
Caching the key-value pairs for the unmasked tokens helps reduce computation, but it can slightly degrade performance. This happens because - (a) when we cache, the unmasked tokens are not updated in the subsequent iterations. (b) when we shift model size during generation, in the attention layer, the query size differs from the cached KV (see \cref{subsec:prelim}). While technically, we can zero-pad the KV to be compatible with the current model's query dimension, the model remains unfamiliar of such feature discrepancies between query and key-value.

To remedy this, we strategically refresh the cache while changing the model size. Refreshing involves discarding the cached KV for that iteration and caching a newly computed KV for the immediate next iteration. We empirically find that it bridges the performance gap that arises due to caching. The proposed decode time model scaling algorithm is presented in \cref{alg:our_algo}, which uses MaskGIT's sampling strategy \cite{chang2022maskgit,yu2023magvit} to sample tokens from logits predicted by the network. 

\begin{figure*}[t]
    \centering
    \begin{minipage}{0.6\textwidth}
            \vspace{0.5cm}
            \resizebox{\textwidth}{!}{ %
            
    \begin{tabular}{lcccccccc}
    \toprule
    Model & AR & FID $\downarrow$ & IS $\uparrow$ & Prec $\uparrow$ & Rec $\uparrow$ & \# params & \# steps & \# Gflops\\
    \midrule
    DCTransformer$^{\square}$ \citep{nash2021generating}&  & 36.5 & - & 36 & 67 & 738M & $\geq$1024 &  -\\
    BigGAN-deep$^{\square}$ \citep{brock2018large}&  & 7.0 & 171.4 & 87 & 28 & 160M & 1 & - \\
    StyleGAN-XL$^{\square g}$ \citep{sauer2022styleganxlscalingstyleganlarge}&  & 2.3 & 265.1 & - & - & 166M & 1 & - \\
    \midrule   %
    Improved DDPM$^{\square}$ \citep{nichol2021improved}&  & 12.3 & - & 70 & 62 & 280M & 250 & $>$5B\\
    ADM + Upsample$^{\square g}$ \citep{dhariwal2021diffusion}&  & 4.6 & 186.7 & 82 & 52 & 554M & 250 &  240k\\
    LDM-4$^{\square g*}$  \citep{ldm}& & 3.6 & 247.7 & - & - & 400M & 250 &  52k\\
    DiT-XL/2$^{\square g*}$ \citep{peebles2023scalablediffusionmodelstransformers}& & 2.3 & 278.2 & 83 & 57 & 675M & 250 & $>$59k \\ %
    Binary latent diffusion$^{\square }$ \citep{wang2023binary}&  & 8.2 & 162.3 & - & - & 172M & 64 & -\\
    MDT$^{\square g* }$ \citep{gao2023masked}&  & 1.8 & 283.0 & 81 & 61 & 676M & 250 & $>$59k \\ %
    MaskDiT$^{\square g* }$ \citep{zheng2023fast}&  & 2.3 & 276.6 & 80 & 61 & 736M & 250 & $>$28k   \\
    CDM$^{\square}$ \citep{ho2022cascaded}&  & 4.9 & 158.7 & - & - & - & 8100 & -  \\
    RIN$^{\square}$ \citep{jabri2022scalable}&  & 3.4 & 182.0 & - & - & 410M & 1000 & 334k  \\
    Simple Diffusion$^{\square g}$ \citep{hoogeboom2023simple}&  & 2.4 & 256.3 & - & - & 2B & 512 & - \\
    VDM++$^{\square g}$  \citep{kingma2023understanding}& & 2.1 & 267.7 & - & - & 2B & 512 & - \\
    MAR$^{\square g}$  \citep{li2024autoregressiveimagegenerationvector}& $\checkmark$ & 1.8 & 296.0 & 81 & 60 & 479M & 128 & - \\
    \midrule
    VQVAE-2$^{\square}$ \citep{razavi2019generating}& $\checkmark$ & 31.1 & $\sim$45 & 36 & 57 & 13.5B & 5120 & -\\
    VQGAN$^{\square}$ \citep{esser2021taming}& $\checkmark$ & 15.8 & 78.3 & - & - & 1.4B & 256 & -\\
    VQGAN (architecture) $+$ MaskGIT (setup)$^{\square}$&  & 18.7 & 80.4 & 78 & 26 & 303M & 256 & -\\
    MaskGIT$^{\square}$\citep{chang2022maskgit}& & 6.2 & 182.1 & 80 & 51 & 303M & 8 & 647 \\
    Mo-VQGAN$^{\square}$ \citep{zheng2022movqmodulatingquantizedvectors}& & 7.2 & 130.1 & 72 & 55 & 389M & 12 & $\sim$1k\\
    MaskBit$^{\square g}$ \cite{weber2024maskbitembeddingfreeimagegeneration}&  & 1.7 & 341.8 & - & - & 305M & 64 & 10.3k\\
    PAR-$4\times^{\square}$ \cite{wang2024parallelizedautoregressivevisualgeneration}&$\checkmark$ & 3.8 & 218.9 & 84 & 50 & 343M & 147 & -\\
    PAR-$16\times^{\square}$ \cite{wang2024parallelizedautoregressivevisualgeneration}&$\checkmark$ & 2.9 & 262.5 & 82 & 56 & 3.1B & 51 & -\\
    \midrule
    MaskGIT++$^{g_{4}}$& & 2.5 & 260.3 & 83 & 54 & 303M & 12 & 1.3k\\ 
    MaskGIT++$^{g_{6}}$& & 2.3 & 280.6 & 84 & 51 & 303M & 16 & 1.8k\\ 
    \textbf{\ours}\ (ours)$^{g_{4}}$& & 3.1 & 254.8 &  85 & 50 & 303M & 12 & 490 \\
    \textbf{\ours}\ (ours)$^{g_{6}}$& & 2.9 & 253.1 &  84 & 51 & 303M & 16 & 608 \\
    \bottomrule
    \end{tabular}

            }
            \captionof{table}{\small  \textbf{Class-conditional Image Generation} on ImageNet $256\times256$. “\# steps” refers to the number of neural network runs. $^{\square}$ denotes values taken from prior publications. $^{*}$ indicates usage of extra training data. $g$ denotes use of classifier-free guidance \citep{ho2022classifierfreediffusionguidance} for all steps. $g_{x}$ represents use of guidance only for final $x$ steps.}
            \label{tab:img_results}
    \end{minipage}
    \hspace{0.2cm} 
    \begin{minipage}{0.35\textwidth}
        \centering
        \includegraphics[trim={2cm 0.5cm 0cm 0cm},width=0.8\textwidth]{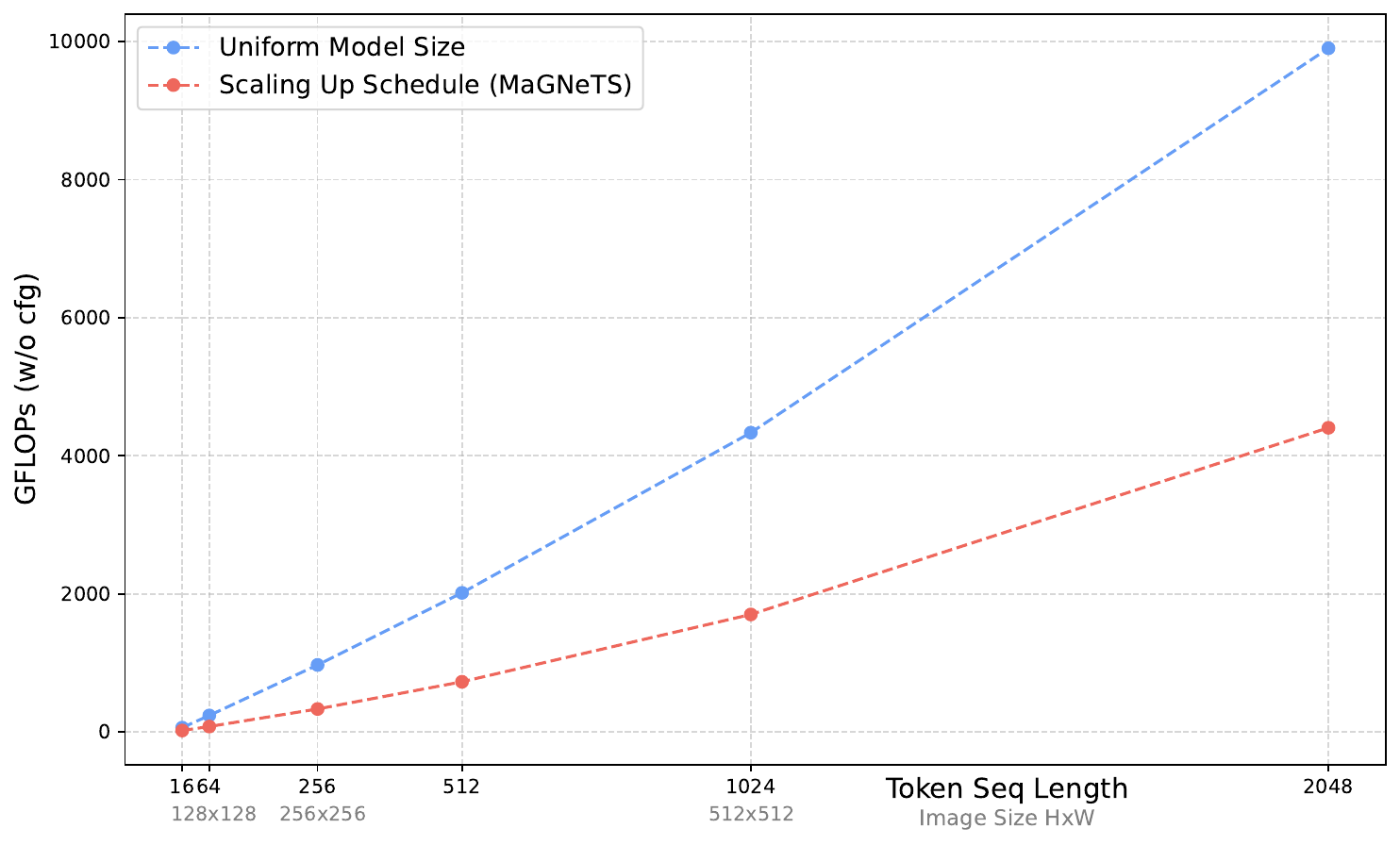}
        \caption{\small \textbf{Compute Comparison} between uniform model schedule (MaskGIT) and \ours, for 12 decode iters.}
        \label{fig:flops_vs_seq}

        \includegraphics[trim={0.5cm 1cm 0cm 0cm},width=\textwidth]{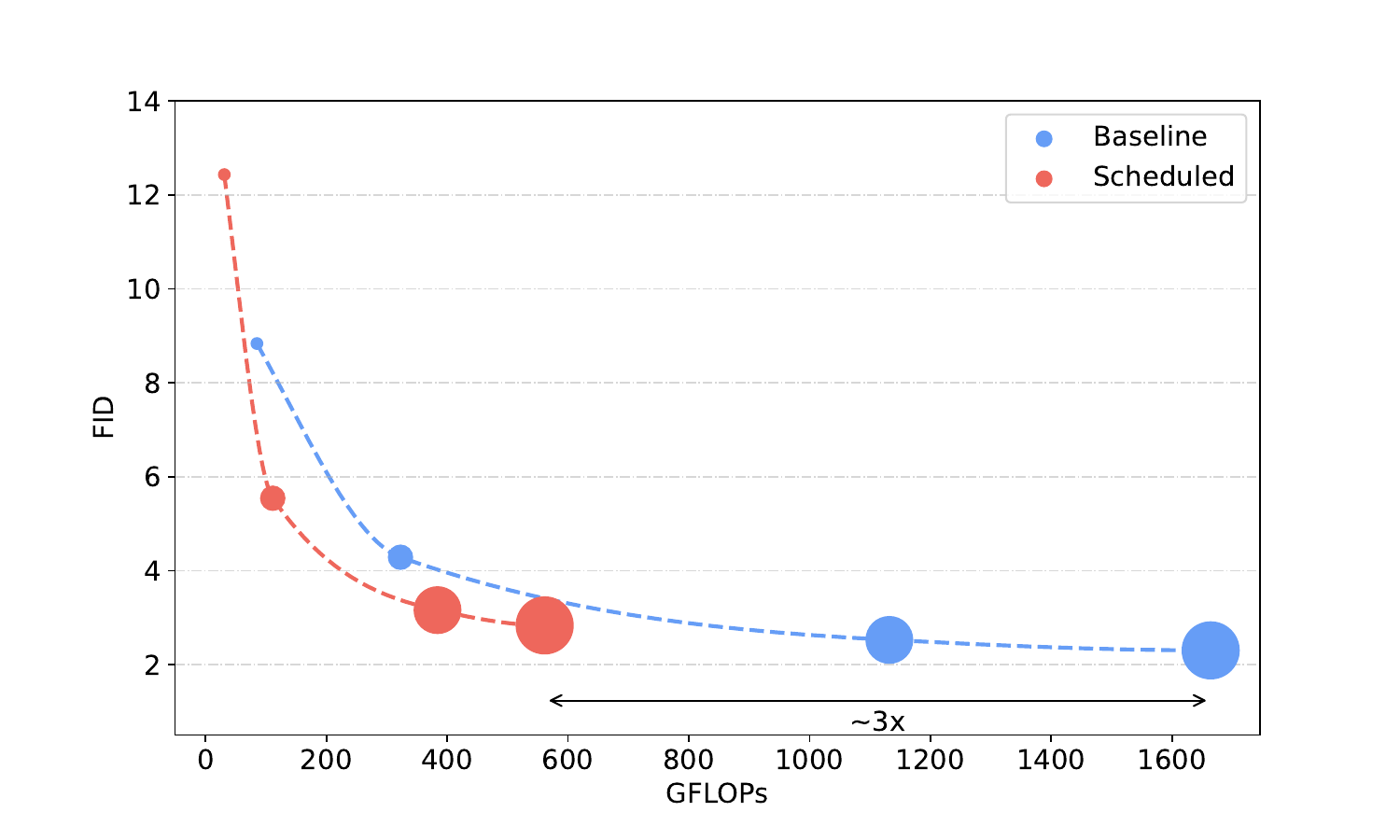}
        \caption{\small \textbf{Compute Scaling Curve.} Generation performance vs compute for different model sizes. The blob size indicates parameter count.}
        \label{fig:scaling_all}
    \end{minipage}
\end{figure*}

{\flushleft \textbf{Training Nested Models.}}
MatFormer \citep{kudugunta2023matformer} opts for a joint optimization of losses w.r.t. ground-truth from all models with equal weights.
While this mode of training works for a small range of model downscaling, we found it to hurt performance with larger downscaling factors $p$. We introduce a combination of ground truth and distillation loss to address this issue. We perform online distillation progressively, where the teacher for  model $m_i$ is model $m_{i+1}$. The full model $m_N (= M)$ is trained with only ground truth loss. This provides a simpler optimization for the smaller nested models while maintaining the overall objective. Progressive distillation also reduces the teacher-student size gap, which can otherwise hurt distillation performance \citep{stanton2021doesknowledgedistillationreally, beyer2022knowledgedistillationgoodteacher,mirzadeh2019improvedknowledgedistillationteacher}. %
Given input $\mathbf{X}$, ground truth label $\mathbf{Y}$ and loss function $\mathcal{L}$, our training loss can be expressed as:
{ \begin{multline}
    \mathcal{L}_{train} = \frac{1}{N} \Big( \mathcal{L}(m_N(\mathbf{X}), \mathbf{Y}) \ + \\ \sum_{i=1}^{N-1} \alpha_i \mathcal{L}(m_i(\mathbf{X}), \mathbf{Y}) + (1-\alpha_i) \mathcal{L}(m_i(\mathbf{X}), m_{i+1}(\mathbf{X})) \Big)
\end{multline}
}
where $\alpha_i$ controls the weight between the distillation and ground truth loss, which is linearly decayed from 1 to 0 as training progresses. Note that a stop gradient is applied during distillation on $m_{i+1}$ in the third term of the equation.

{\flushleft \textbf{Classifier-Free Guidance.}} Following literature \cite{ho2022classifierfreediffusionguidance,yu2023language}, we also utilize classifier-free guidance during the generation process. Following the same motivation as decode time model scaling discussed above, which shows that the initial decoding iterations focus on the background region, and gradually moves to the main object/region of interest in the final decoding iterations, we experiment with applying guidance to only to a few final decoding iterations. We find that adding guidance only to few final iterations offers similar quality images as applying to all (refer \cref{fig:guid}). See \cref{sec:sample} for detailed analysis.

\section{Experiments and Results}
We conduct extensive experiments to demonstrate the efficacy of our approach on three distinct tasks: class-conditional image generation, class-conditional video generation, and frame prediction.

{\flushleft \textbf{Datasets.}}
We evaluate our model on ImageNet 256 $\times$ 256 \citep{deng2009imagenet} for image generation, UCF101 \citep{soomro2012ucf101dataset101human} for video generation and Kinetics600 \citep{carreira2018shortnotekinetics600} for frame prediction (5-frame condition). %

{\flushleft \textbf{Implementation Details.}} We utilize the pretrained tokenizers from MaskGIT \citep{chang2022maskgit} (for images) and MAGVIT \citep{yu2023magvit} (for videos)
with the codebook size of 1024 tokens.
We train models for image size $256 \times 256$. The tokenizer compresses it to $16 \times 16$ discrete tokens. For videos, we learn models for $16 \times 128 \times 128$, where the tokenizer outputs $4 \times 16 \times 16$ tokens. Following MaskGIT, we utilize the Bert model \citep{devlin2019bertpretrainingdeepbidirectional} as a transformer backbone. We perform experiments at several model scales to understand the scaling behaviors of our algorithm. We utilize the same training hyper-parameters to
train our nested models as these baselines. We train our model for 270 epochs for all the experiments.
Unless otherwise mentioned, throughout the paper, we employ same number of steps per model before switching to the next model, i.e., $k_{1} = k_{2} = .... = k_{n}$. We follow a cosine schedule of unmasking tokens during inference. For image generation and frame prediction, we use classifier-free guidance for both \ours \ and respective baselines. Following literature, we drop input class condition labels for $10\%$ of the training batches in image generation. We mention the details of sampling hyperparameters in \cref{sec:sample}.

{\flushleft \textbf{Evaluation Metrics.}}
Following previous baselines, we use Fréchet Inception Distance (FID) \citep{heusel2017gans, dhariwal2021diffusion} for image generation, Fréchet Video Distance (FVD) \citep{unterthiner2019fvd} for the video generation tasks, Inception Score \citep{salimans2016improvedtechniquestraininggans} for both tasks, as well as precision and recall for image generation. We compare algorithms using inference-time GFLOPs. Refer \cref{sec:compute_gains} for GFLOPs computation details.

\subsection{Image Generation}

{\flushleft \textbf{Comparison with Baselines.}} In this section, we compare \ours \ with state-of-the-art methods in the literature for image generation. We list the results in \cref{tab:img_results} for $256 \times 256$ image generation on ImageNet-1k. 
\cref{tab:img_results} shows that \ours \ can speed up the generation process by $2.65 - 3\times$ (depending on total step count), with a negligible drop in FID. Refer \cref{sec:compute_gains} for real-time gains. \cref{fig:flops_vs_seq} illustrates that \ours \ significantly accelerates parallel decoding, which gets more pronounced as image resolution grows. \cref{fig:qual_img} and \cref{fig:qual_img2} show generated images from MaskGIT++ and \ours \ (ours). 
As shown before in \citet{yu2023language, weber2024maskbitembeddingfreeimagegeneration}, using a superior tokenizer can further boost MaGNeTS's performance. Note that several recent diffusion-based works only report results on the low-resolution of ImageNet (typically 64$\times$64), and therefore a direct comparison is not possible.

{\flushleft \textbf{Scaling Analysis.}} To understand the scaling properties of \ours \, we train models of different sizes - S (22M), B (86M), L (303M) and XL (450M) for both the baseline as well as nested models needed for our algorithm. We use the same hyper-parameters for all, such as learning rate, epochs, weight decay, etc. We present the results in \cref{fig:scaling_all}. It shows the compute vs performance of different models, with the blob size denoting the model size. For a certain parameter count, the baseline uses the full model for all $12$ decoding steps, while the scheduled routines use a sequence of nested models with downsampling factors  $p=8, 4, 2, 1$ for $3$ steps each. It can be seen that scaling up model size lead to much cheaper compute scaling of \ours \ than the baseline, with almost $3\times$ compute reduction. %

\subsection{Video Generation}
We use the MAGVIT \citep{yu2023magvit} framework to train parallel decoding based video generation and frame prediction models. \cref{fig:ucfqual} shows generated videos of UCF101. We summarize the results for class-conditional video generation on UCF101 in \cref{tab:ucf101_results} and for frame prediction on Kinetics600 in \cref{tab:k600_quant}. Despite the challenging nature of video generation relative to image generation, results indicate that the decode time scaling of model size holds true even for video generation. \ours \; remains competitive to MAGVIT for frame prediction with $\sim 3.7 \times$ lower compute.\begin{table}[!t]
    \centering
    \resizebox{0.48\textwidth}{!}
    {
        \begin{tabular}{lcccccc}
            \toprule
            Method & Class & FVD $\downarrow$ & IS$\uparrow$  &  \# params & \# steps & \# GFlops\\
            \midrule
            RaMViD$^{\square*}$ \citep{hoppe2022diffusion}  &  & - & 21.71 $\pm$  \textcolor{gray}{0.21} & 308M & 500 & -\\
            StyleGAN-V$^{\square*}$\citep{skorokhodov2022stylegan}  &  & - & 23.94 $\pm$  \textcolor{gray}{0.73} & - & 1 & -\\
            DIGAN$^{\square}$ \citep{yu2022generating} &  & 577$\pm$\textcolor{gray}{21} & 32.70$\pm$ \textcolor{gray}{0.35} & - & 1 & $\sim$148\\
            DVD-GAN$^{\square}$ \citep{clark2019adversarial}  & $\checkmark$ & - & 32.97$\pm$ \textcolor{gray}{1.70} & - & 1 & -\\
            Video Diffusion$^{\square*}$ \citep{ho2022video}  &  &  & 57.00$\pm$ \textcolor{gray}{0.62} & 1.1B & 256 & -\\
            TATS$^{\square}$ \citep{ge2022longvideogenerationtimeagnostic} &  & 420$\pm$ \textcolor{gray}{18} & 57.63$\pm$ \textcolor{gray}{0.24} & 321M & 1024 & -\\
            CCVS+StyleGAN$^{\square}$ \citep{le2021ccvs} &  & 386$\pm$ \textcolor{gray}{15} & 24.47$\pm$ \textcolor{gray}{0.13}& - & - & - \\
            Make-A-Video$^{\square*}$ \citep{singer2022make}  &  $\checkmark$& 367 & 33.00 & - & - & - \\
            TATS$^{\square}$ \citep{ge2022longvideogenerationtimeagnostic} & $\checkmark$  & 332$\pm$ \textcolor{gray}{18} & 79.28$\pm$ \textcolor{gray}{0.38} & 321M & 1024 & - \\
            \midrule
            \textcolor{gray}{CogVideo}$^{\square*}$ \citep{hong2022cogvideo} & $\checkmark$ & 626 & 50.46 & 9.4B & - & - \\
            \textcolor{gray}{Make-A-Video}$^{\square*}$ \citep{singer2022make} & $\checkmark$  & 81 & 82.55 & $\gg$3.5B & $\gg$250 & -\\
            PAR-4$\times^{\square}$ \cite{wang2024parallelizedautoregressivevisualgeneration} &$\checkmark$ & 99.5 & - & 792M & 323 & -\\
            PAR-16$\times^{\square}$ \cite{wang2024parallelizedautoregressivevisualgeneration} &$\checkmark$ & 103.4 & - & 792M & 95 & -\\
            \midrule
            MAGVIT-B$^{\square}$ \citep{yu2023magvit} & $\checkmark$ & 159$\pm$ \textcolor{gray}{2} & 83.55$\pm$ \textcolor{gray}{0.14} & 87M &12 & $\sim$1.3k\\ %
            MAGVIT-L \citep{yu2023magvit} & $\checkmark$ & 74.4$\pm$ \textcolor{gray}{2} & 89.54$\pm$ \textcolor{gray}{0.21}  & 306M & 12 & $\sim$4.3k \\ %
            \textbf{\ours} (ours) & $\checkmark$ &96.4$\pm$\textcolor{gray}{2} &  88.53$\pm$\textcolor{gray}{0.20} & 306M & 12 & $\sim$1.7k \\ %
            \bottomrule
        \end{tabular}
     }
     \caption{\small \textbf{Class-conditional Video Generation} on UCF-101. Methods in  \textcolor{gray}{gray} are pretrained on additional large video data. Methods with $\checkmark$ in the Class column are class-conditional, while the others are unconditional. Methods marked with $^{*}$ use custom resolutions, while the others are at 128$\times$128. $^\square$ denotes values taken from prior publications. No guidance is used for UCF101.}
     \label{tab:ucf101_results}
     \vspace{-5mm}
\end{table}

\begin{table}[h]
    \centering
    \resizebox{0.49\textwidth}{!}{
    \begin{tabular}{lccccc}
        \toprule
        Method & FVD $\downarrow$ & IS $\uparrow$ & \# params & \# steps & \# GFlops\\
        \midrule
        CogVideo$^{\square}$ \citep{hong2022cogvideo} & 109.2 & - &9.4B & - & -\\
        CCVS$^{\square}$  \citep{le2021ccvs} & 55.0$\pm$\textcolor{gray}{1.0} & - & - & - & -\\
        Phenaki$^{\square}$  \citep{villegas2022phenaki} & 36.4 $\pm$ \textcolor{gray}{0.2} & - & 1.8B & 48 & -\\
        TrIVD-GAN-FP$^{\square}$  \citep{luc2020transformation} & 25.7 $\pm$ \textcolor{gray}{0.7} & 12.54 $\pm$ \textcolor{gray}{0.06} & - & 1 & - \\
        Transframer$^{\square}$  \citep{nash2022transframer} & 25.4 & - & 662M & - & -\\
        RaMViD$^{\square}$  \citep{hoppe2022diffusion} & 16.5 & - &308M & 500 & - \\
        Video Diffusion$^{\square}$ \citep{ho2022video} & 16.2 $\pm$ \textcolor{gray}{0.3} & 15.64 & 1.1B & 128 & -\\
        \midrule
        MAGVIT-B$^{\square}$ & 24.5$ \pm$ \textcolor{gray}{0.9} & - & 87M & 12 & $\sim$1.3k\\
        
         MAGVIT-L& 7.2 $\pm$ \textcolor{gray}{0.1} & 16.48 $\pm$ \textcolor{gray}{0.01} &306M &12 & $\sim$ 4.3k\\
         
        MAGVIT-L$^{g_{2}}$& 6.6 $\pm$ \textcolor{gray}{0.1} & 16.29 $\pm$ \textcolor{gray}{0.01} &306M &12 & $\sim$ 5.1k\\
        \hdashline
        \textbf{\ours} (ours) &  10.8 $\pm$ \textcolor{gray}{0.1} &  16.25 $\pm$ \textcolor{gray}{0.02} & 306M & 12 & $\sim$1.2k \\ 
        
        \textbf{\ours} (ours)$^{g_{2}}$ &  9.6 $\pm$ \textcolor{gray}{0.1} &  16.25 $\pm$ \textcolor{gray}{0.01} & 306M & 12 & $\sim$1.4k \\ 
        \bottomrule
    \end{tabular}
    }
    \caption{\small \textbf{Frame prediction} on K600. $^\square$ denotes values taken from papers. $g_{x}$ denotes use of guidance only for final $x$ steps.}
    \label{tab:k600_quant}
    \vspace{-2mm}
\end{table}

\begin{figure*}[t!]
    \centering
    \begin{subfigure}[b]{0.45\textwidth}
        \includegraphics[width=\textwidth]{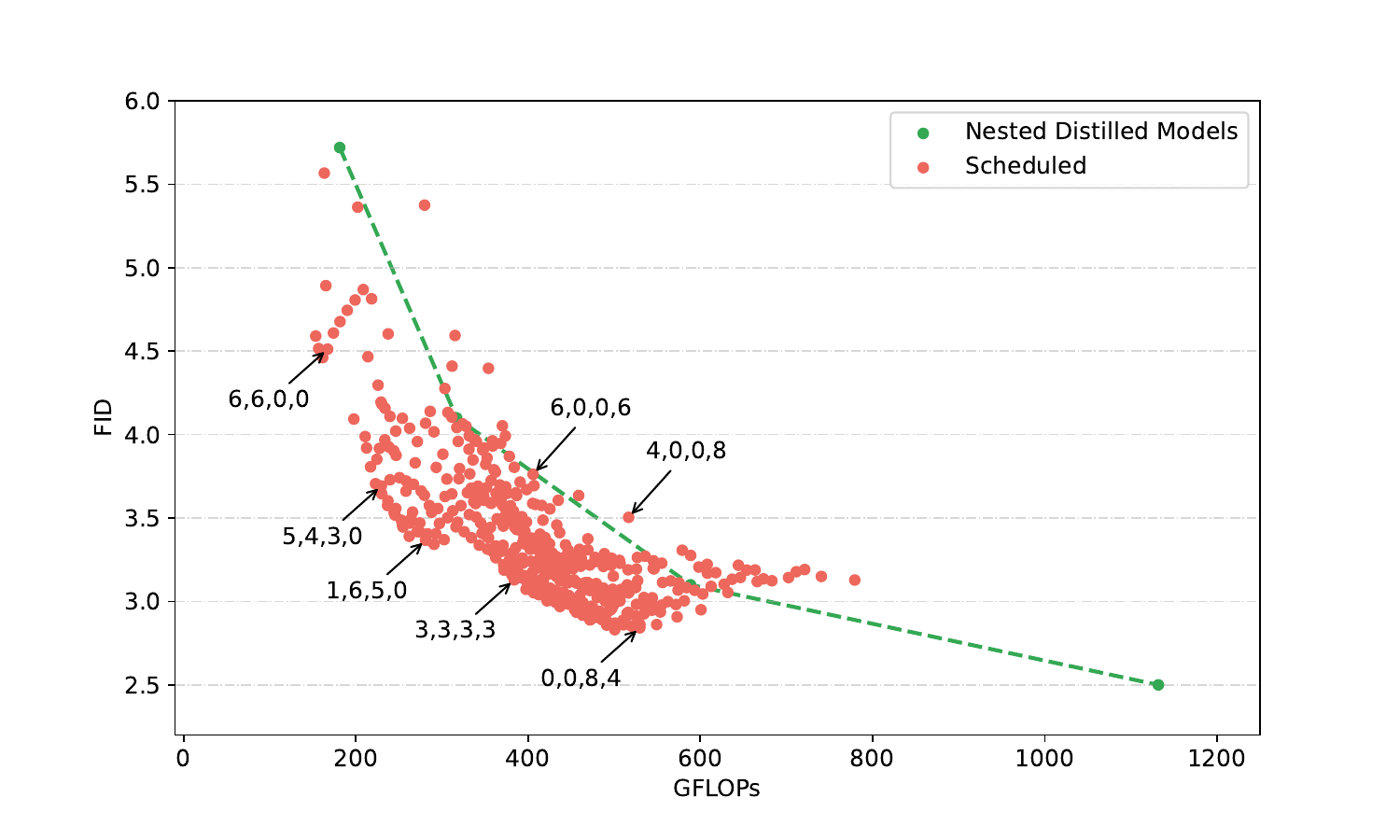}
        \caption{Scaling Up Schedules}
        \label{fig:scaling_up}
    \end{subfigure}
    \begin{subfigure}[b]{0.45\textwidth}
        \includegraphics[width=\textwidth]{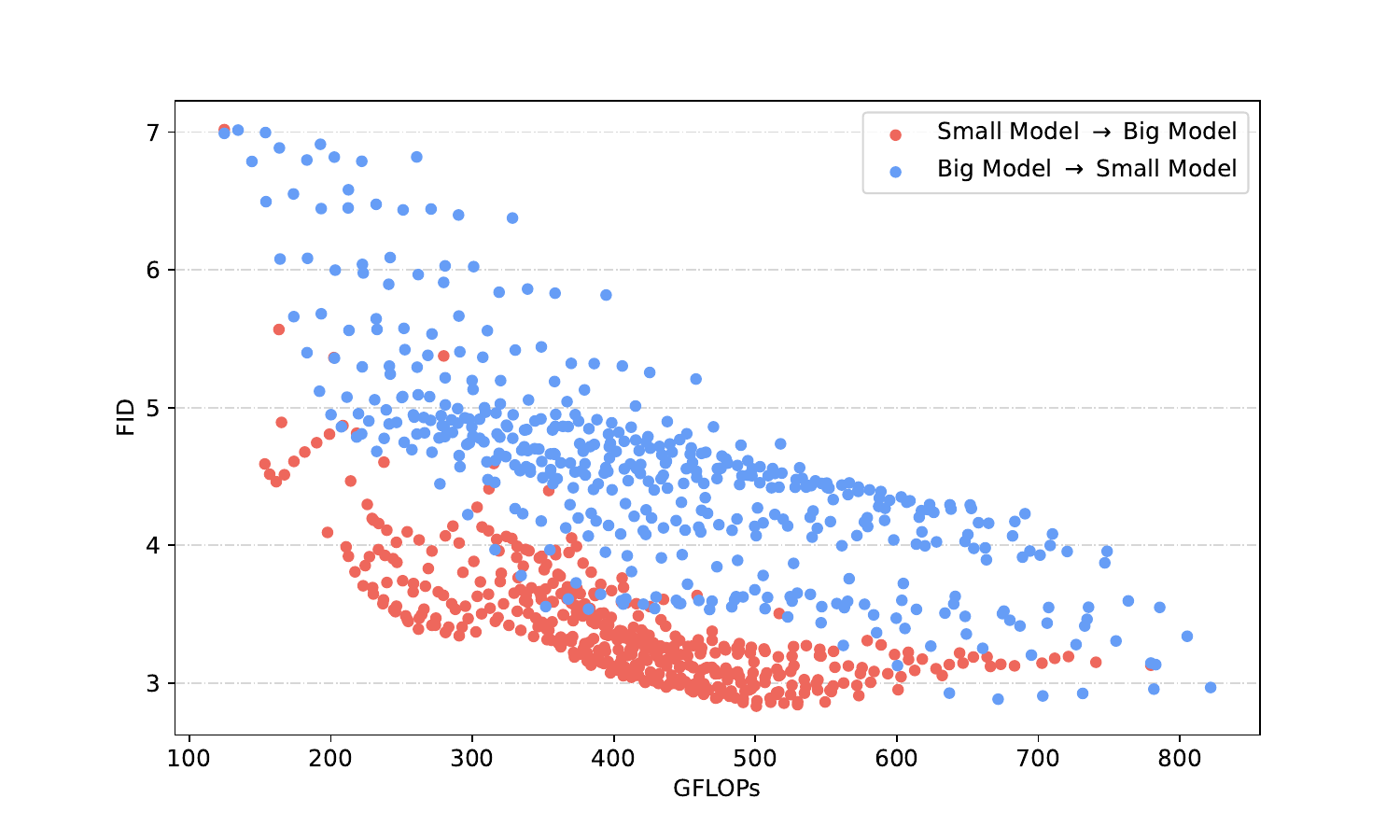}
        \caption{Scaling up vs down}
        \label{fig:scale_down_up}
    \end{subfigure}
    \caption{\small \textbf{Scheduling Options.} (a) This shows the compute-performance trade-off for different schedule options while always scaling up model size over generation iters. The four numbers for each point denote the number of iters each model size operates in the order of downsampling factor $p=(8,4,2,1)$. (b) This shows the benefit of scaling up model size compared to scaling it down during decoding.}
    \label{fig:schedule}
\end{figure*}

\subsection{Ablation Studies}

{\flushleft \textbf{Impact of Decode Time Model Schedule.}} We study the effect of different model scheduling choices. As discussed previously, we can think of different model schedules - scaling up model size, scaling down, periodic scaling up and down, and so on. For this analysis, we consider the L-sized model, with three nested models within it with parameter reduction by roughly $\frac{1}{2}, \frac{1}{4}, \frac{1}{8}$. We can denote the number of times these four models are called during decoding as $(k_1, k_2, k_3, k_4), \text{s.t.}, \sum_{i=1}^4 k_i = 12$. We drop the model notation of $m_p$ in \cref{eqn:schedule} for simplicity and explicitly mention the model names in the text, as discussed next.

First, we evaluate all combinations of $k_i$ for which we always scale up in \cref{fig:scaling_up} in red and scale down in \cref{fig:scale_down_up} in blue. The green curve shows the performance of the individual nested models. We have the following observations - (1) for a certain compute budget, the scheduling of models over generation iterations (red dots) can offer better performance than using a single nested models (green curve) for all steps. (2) Models that have smoother transitions in nested models, such as $(3, 3, 3, 3)$ or $(0, 0, 8, 4)$, offer much better performance than the ones which has abrupt model transition such as $(6, 0, 0, 6)$ or $(3, 0, 0, 9)$, i.e., directly jumping from the smallest to the biggest model. (3) \cref{fig:scale_down_up} shows that scaling up nested model size offers much better performance than scaling down model size. This shows that bigger models are better utilized in the later iters.

{\flushleft \textbf{Impact of Caching and Refresh.}}
We now discuss the impact of caching and its refresh. For this analysis, we use a uniform model schedule: $k_1=k_2=k_3=k_4=3$. We also perform caching and refresh on the baseline model, which has not been trained with any nesting and has the same model applied for all iterations.
We also refresh the cache at exactly the same steps as the scheduled model for the baseline. We present the results in \cref{tab:cache_ablation}. The columns ``Baseline" and ``Scheduled" do not involve any cache. While caching degrades the performance a bit, refreshing it intermittently can avoid the degradation. While refresh does have some compute overhead, it does help significantly. Scheduling of models with caching and refresh has the best compute-performance trade-off. 

\begin{table}[h]
\small
\centering
\resizebox{0.49\textwidth}{!}{
\begin{tabular}{l| ccc | ccc}
\toprule
Algorithm & Baseline & + Cache & + Refresh & Scheduled & + Cache & + Refresh \\ 
\midrule
FID  & 2.5 & 3.4 & 2.6 & 3.1 & 4.8 & 3.1  \\
FLOP Gains (times) & 1.0 & 1.3 & 1.2 & 2.1 & 3.5 & 3.0\\
\bottomrule
\end{tabular}
}
\caption{\small \textbf{Caching Ablation.}  As we can see, adding caching does take a hit in performance, which can be regained by cache refresh. Scheduling of models (scaling up) with caching and its intermittent refresh offers the best compute-performance trade-off. These results are on ImageNet256$\times$256 with model size L.}
\label{tab:cache_ablation}
\end{table}

{\flushleft \textbf{The efficiency of using nested models.}}
In \ours \ we use nested models instead of separately trained smaller sized models. This has two advantages - (a) parameter sharing, which limits the number of parameters to just that of the full model, compared to $1.875\times (=1+1/2+1/4+1/8)$ for disjoint models. Increasing the parameter count will increase memory requirements. (b) Nested models can be trained efficiently in just a single training run. When trained with distillation, they generate better models than training standalone models (refer \cref{sec:ablation_supp}) of the same size as the nested models. For performance comparison, we trained standalone (L-sized) models of the same size as the nested models for both UCF101 and ImageNet. The results are presented in \cref{tab:std_vs_nested}. Nested models can efficiently share parameters without loss in performance (ImageNet) and offer constraints that help in better performance (UCF101) than using standalone models.

\begin{table}[h]
\small
\centering
\begin{center}
\resizebox{0.4\textwidth}{!}{
\begin{tabular}{l| cc}
\toprule
Dataset & Nested Models & Standalone Models \\ 
\midrule
ImageNet (FID) & 3.1 & 3.1 \\
UCF101 (FVD) & 96.4 & 115.0 \\
\bottomrule
\end{tabular}
}
\caption{\small \textbf{Nested vs Standalone Models.} This table presents the performance comparison between using nested models vs. standalone, independent models without parameter sharing in the decode-time scheduling algorithm of \ours.}
\label{tab:std_vs_nested}
\vspace{-5mm}
\end{center}
\end{table}

\section{Conclusion}
In this paper, we propose \ours, a novel approach for allocating different compute to different steps of the image/video generation process. We show that 
instead of always using the same sized transformer model for all decoding steps, we can start from a model which is nested and fraction of its full size, and then gradually increase model size. This along with key-value caching in the parallel decoding paradigm obtains significant compute gains. We believe that our exploration of dynamic compute opens exciting new directions for research in efficient and scalable generative models. In future works, we plan to explore token-dependent model schedules for further compute gains.

\nocite{langley00}

\bibliography{main}
\bibliographystyle{icml2025}

\newpage
\mbox{}
\newpage
\appendix
\begin{figure*}[!h]
    \centering
    \includegraphics[trim={1cm 0cm 2cm 0cm},   width=0.95\textwidth]{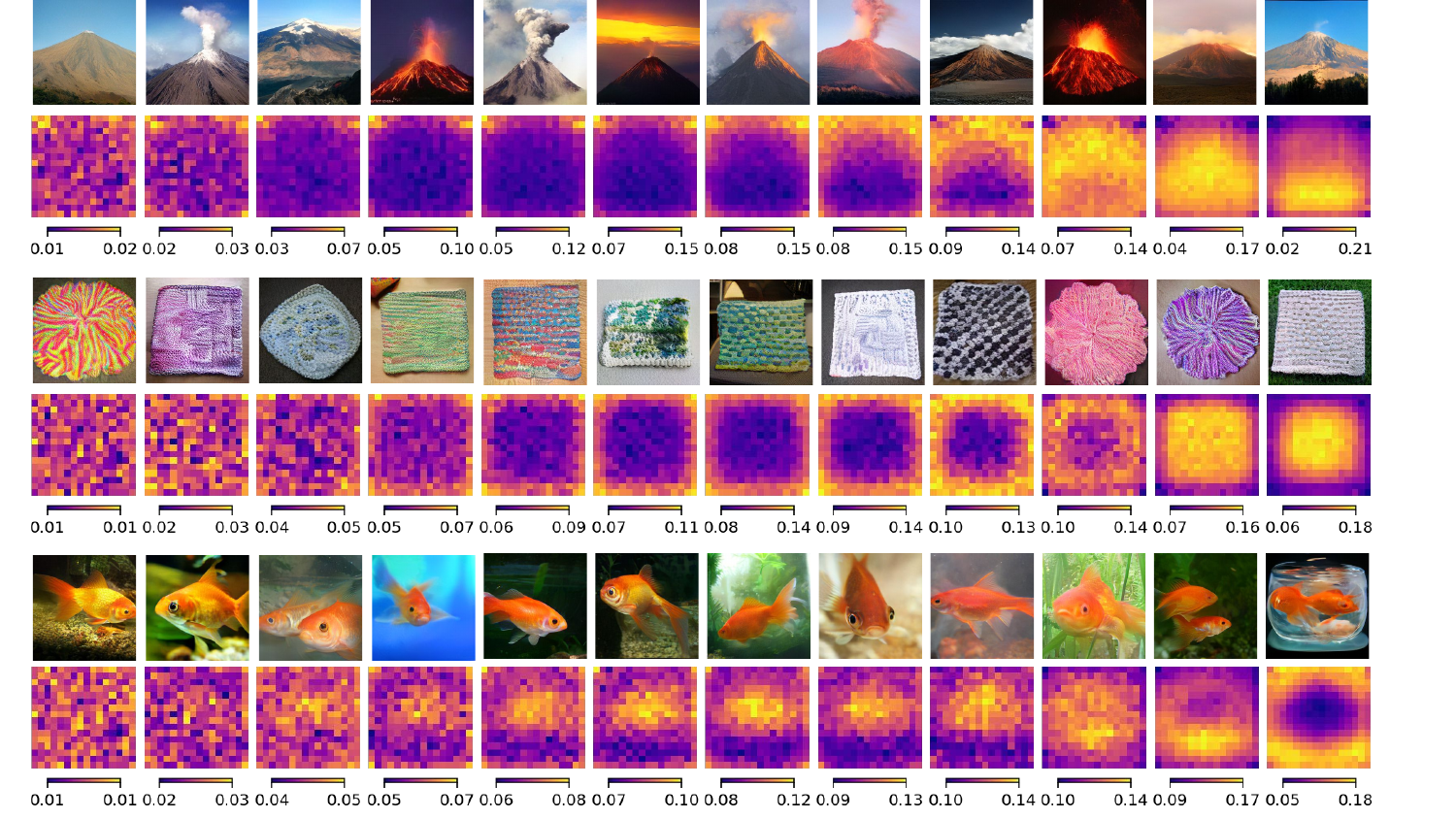}
    \caption{Visualization of token density unmasked in each iteration averaged over 10k generated samples on different categories of ImageNet. The top example shows category \textit{volcano} (non-center-focused). Middle and bottom examples show \textit{dishrag,dishcloth} and \textit{goldfish,Carassius auratus} center-focused categories, respectively.
    Yellow color represents higher density, and each pixel represents a token from the $16 \times 16$ token space.}
    \label{fig:vis_tokens_supp}
\end{figure*}

\begin{figure*}[t]
    \centering
    \begin{subfigure}[b]{0.45\textwidth}
        \centering
        \includegraphics[scale=0.33]{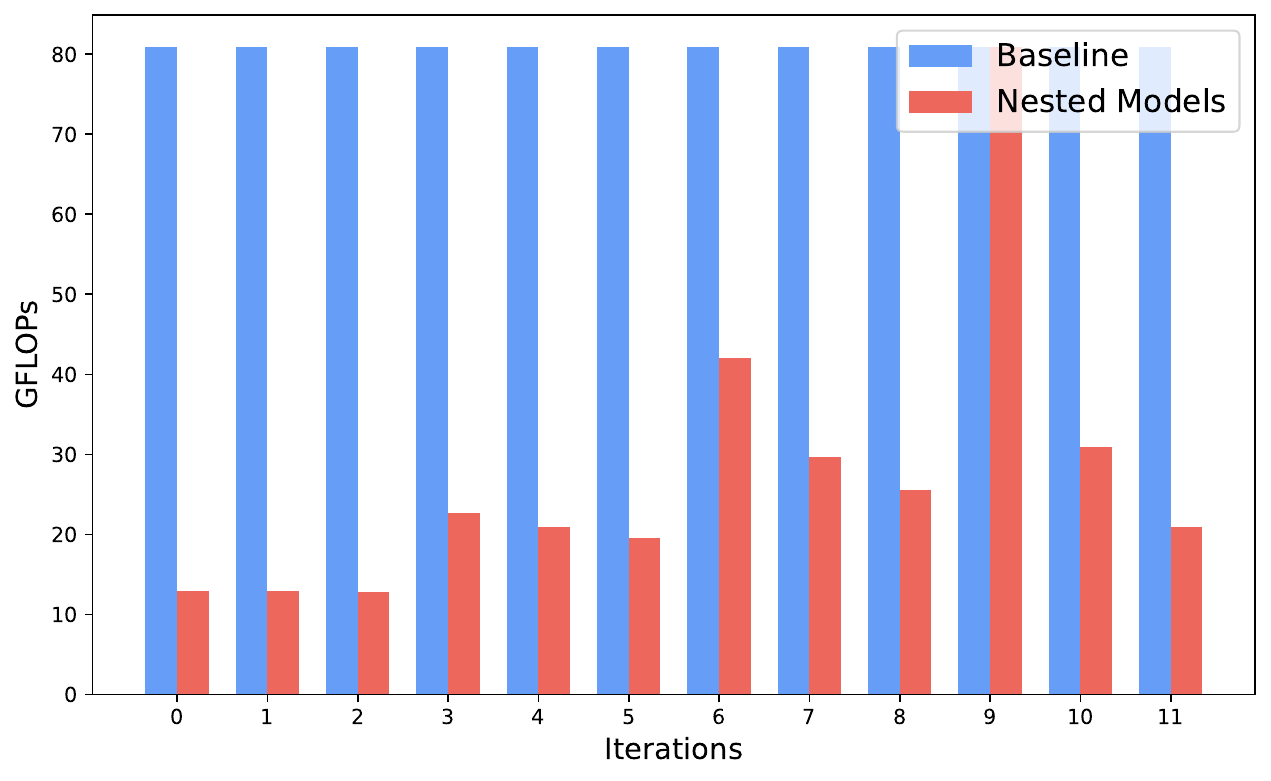}
        \caption{\small GFLOPs per sampling step.}
        \label{fig:flops_over_iter}
    \end{subfigure}
    \hfill
    \begin{subfigure}[b]{0.48\textwidth}
        \centering
        \includegraphics[trim={1cm 0cm 2cm 0cm}, width=\textwidth]{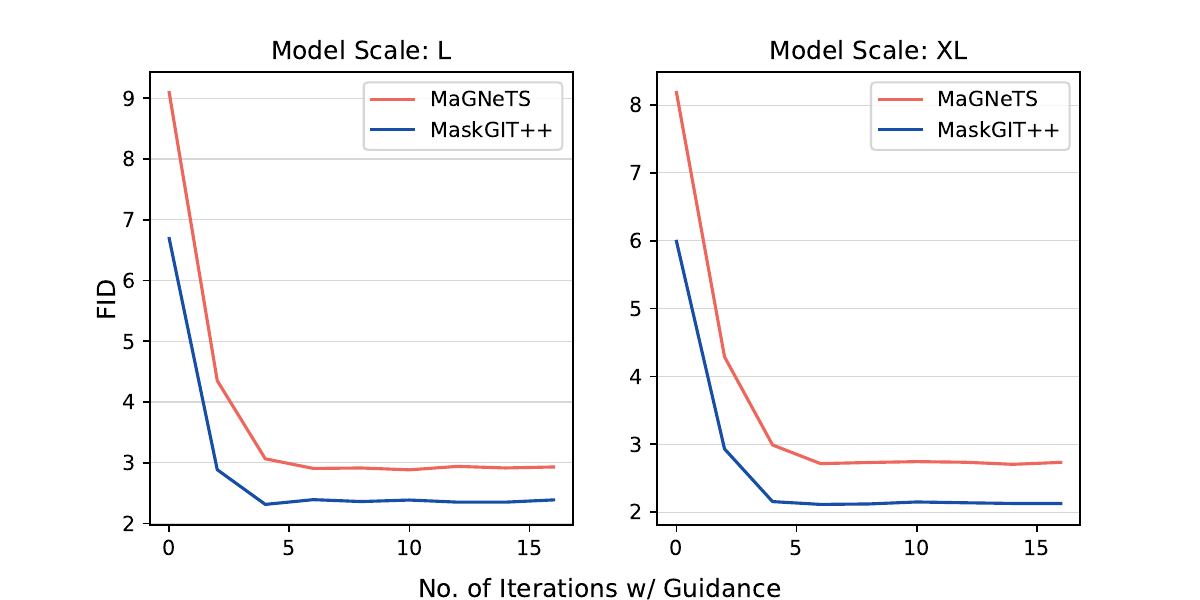}
        \caption{\small Guidance Analysis.}
        \label{fig:guid}
    \end{subfigure}
    \caption{\small (a) Inference GFLOPs per step for baseline and \ours. (b) generation performance (FID) on ImageNet vs Number of decoding iterations w/ guidance for different model scales. Note that we start from last decoding iteration. For example, "No. of iterations w/ Guidance $= 6$" means we use guidance only for final six iterations (out of total 16 iterations). This shows that using guidance only for few final iterations is enough in the parallel decoding setup.}
    \label{fig:combined_fig}
\end{figure*}

\begin{figure*}[ht]
    \centering
    \includegraphics[width=0.95\textwidth]{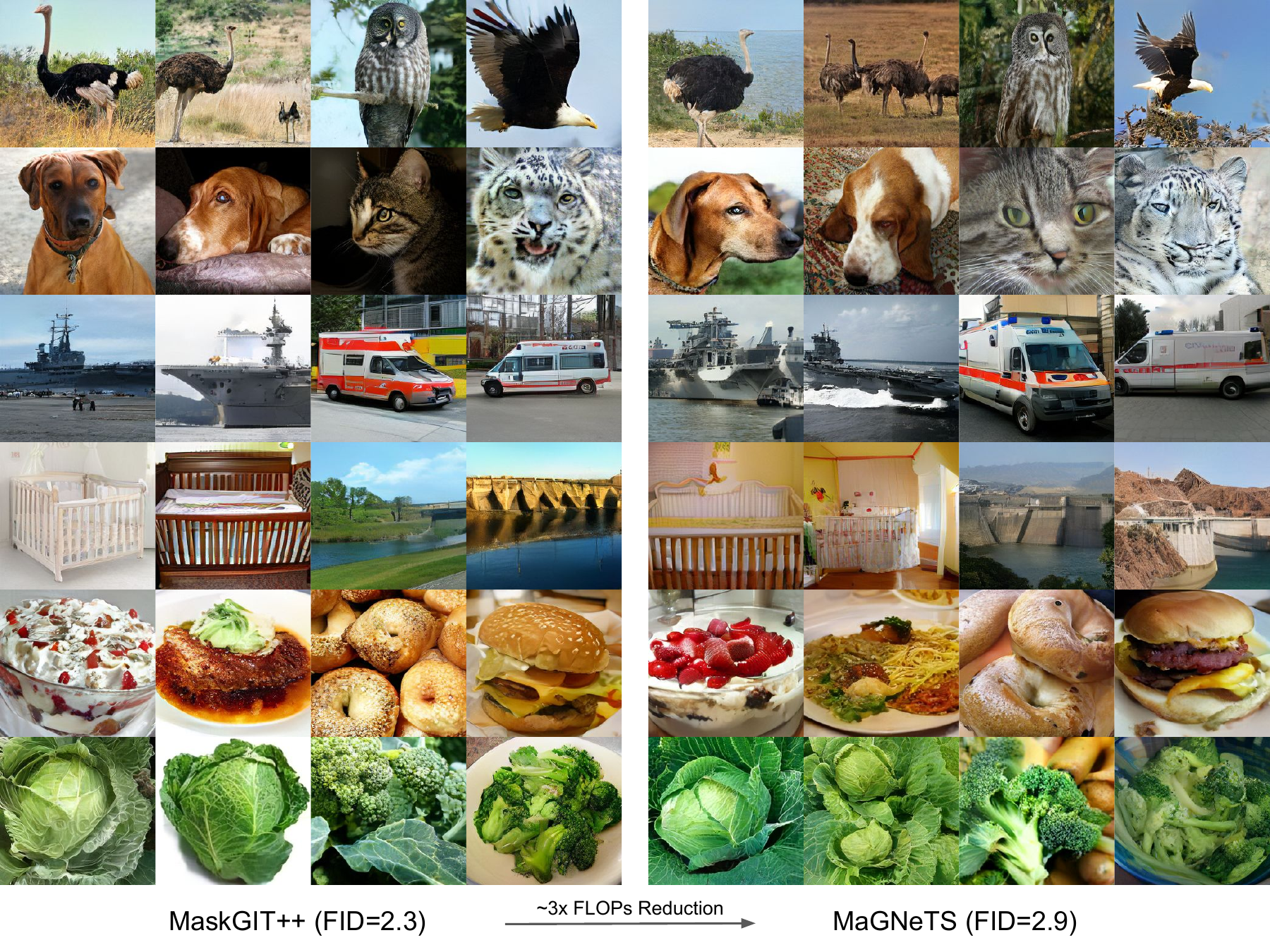}
    \caption{\textbf{Class-conditional Image Generation.} More qualitative results on ImageNet. Comparing MaskGIT++ and \ours \ (size: L, epochs: 270). }
    \label{fig:qual_img2}
\end{figure*}

\begin{figure*}[ht]
    \centering
    \includegraphics[trim={1.5cm 4.5cm 11.5cm 2cm}, width=0.95\textwidth]{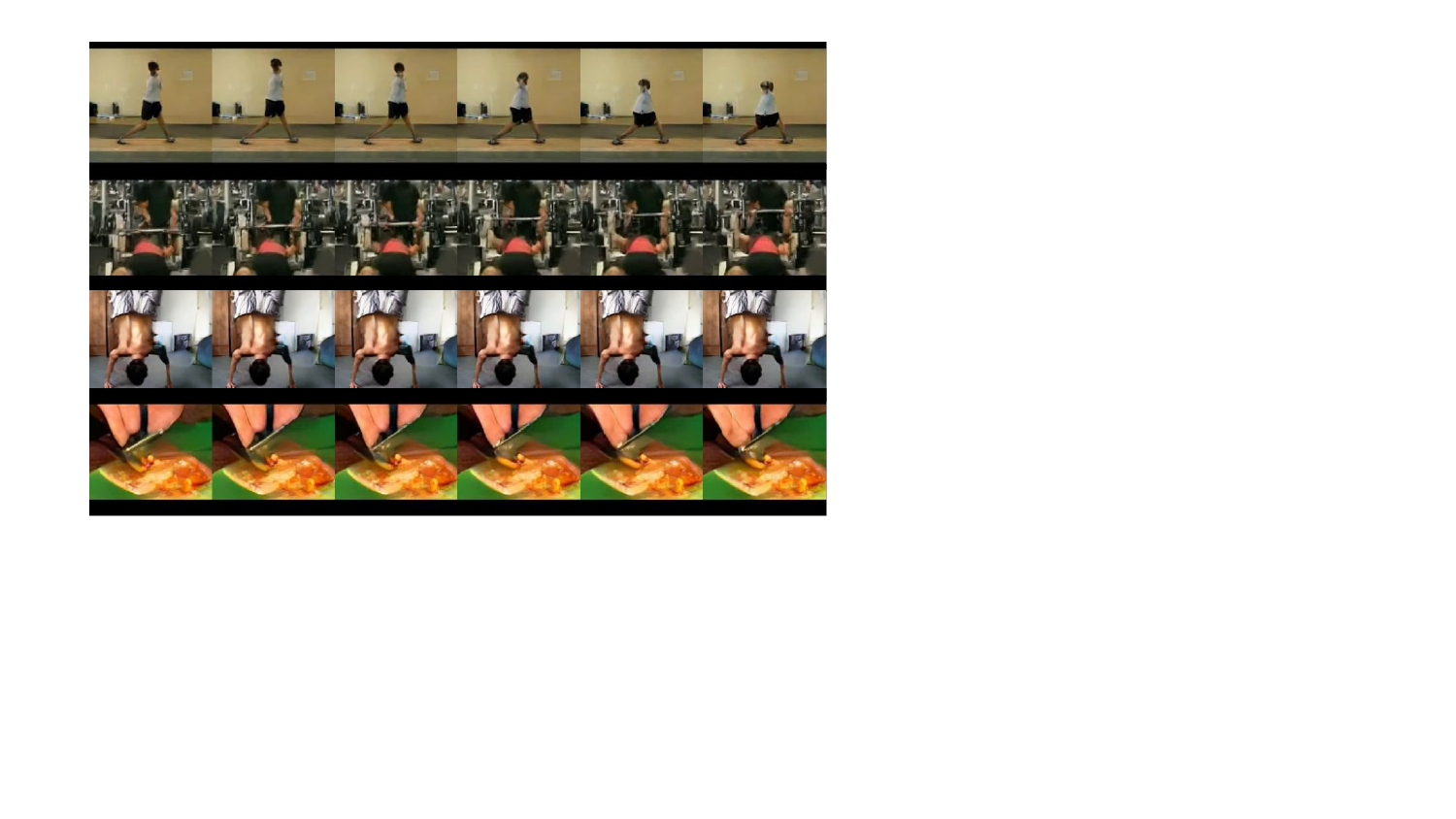}
    \caption{\textbf{Class-conditional Video Generation on UCF101.} 16-frame videos are generated at 128×128 resolution 25 fps. Every third frame is shown for each video. The classes from top to bottom are \textit{Lunges}, \textit{Bench Press}, \textit{Handstand Pushups}, \textit{Cutting In Kitchen}.}
    \label{fig:ucfqual}
\end{figure*}

\begin{figure*}[h]
    \centering
    \includegraphics[trim={0cm 14cm 0cm 0cm}, width=0.95\textwidth]{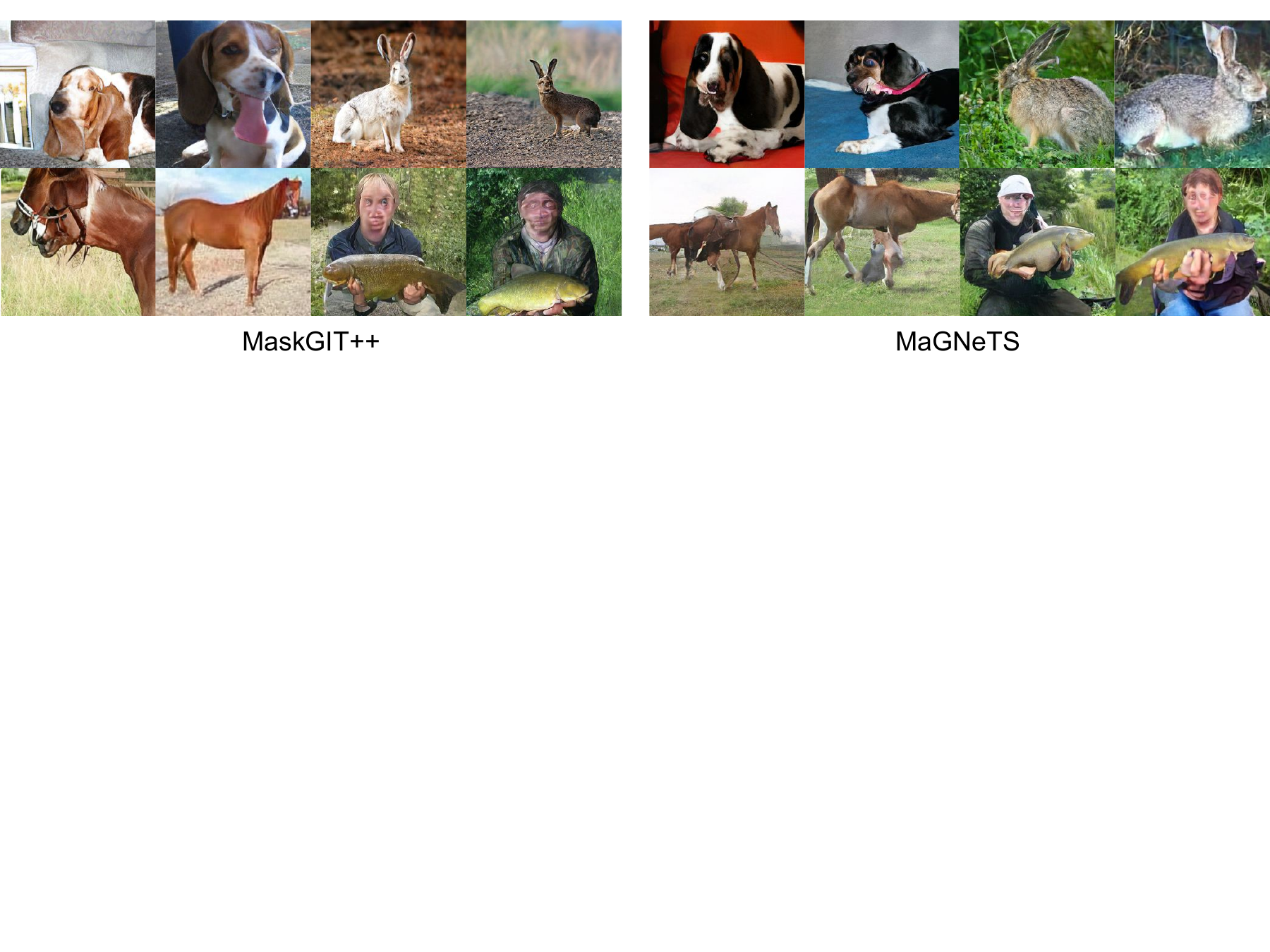}
    \caption{\textbf{Failure cases.} Similar to existing methods, our system can produce results with noticeable artifacts.}
    \label{fig:fail}
\end{figure*}

\section{Motivation for Decode Time Model Scaling} \label{appendix:motivation}
Our visualization of token density averaged across 50k ImageNet samples reveals a dynamic pattern - initial decoding iterations prioritize background regions. In contrast, later iterations focus on the center where foreground objects or region of interest typically reside. This highlights the need to allocate resources efficiently during generation. To further investigate this behavior, we examine token density across various ImageNet categories (refer \cref{fig:vis_tokens_supp}). This category-wise analysis further motivates our focus on decode time scaling. \cref{fig:qual_img2} shows more qualitative results on ImageNet$256\times256$ and \cref{fig:ucfqual} shows samples on UCF101.

\section{Hyper-parameter Details}
\label{sec:sample}

The MaskGIT algorithm has the following hyper-parameters which we discuss next.

{\flushleft \textbf{Guidance Scale ($\mathbf{gs}$).}} It is used in classifier-free guidance \citep{ho2022classifierfreediffusionguidance} and governs the calculation of final logits during inference as shown in \cref{eq:gs}.
\begin{equation}
\mathrm{logits}_\mathrm{final} = \mathrm{logits}_\mathrm{cond} + \lambda \cdot \mathrm{gs} \cdot (\mathrm{logits}_\mathrm{cond} - \mathrm{logits}_\mathrm{uncond})
\label{eq:gs}
\end{equation}
where $\mathrm{logits}_\mathrm{cond}$ are from class-conditional input, $\mathrm{logits}_\mathrm{uncond}$ are from unconditional input, and $\lambda$ depends on the mask-ratio of the current decoding iteration. 

\cref{fig:vis_tokens_supp} shows that the initial decoding iterations of parallel decoding focus on the background region, and focus gradually shifts to the main object/region in the final decoding iterations. Motivated by this, we experimented with applying guidance to only few final decoding iterations and present our findings in \cref{fig:guid}. As we can see, most of the decoding iterations do not require guidance. We use guidance only for final few decoding iterations for class-conditional generation in ImageNet256$\times$256 and frame prediction in Kinetics600. Following MAGVIT \cite{yu2023magvit}, for class-conditional generation in UCF101 we do not use classifier-free guidance.

\paragraph{Mask Temperature ($\mathbf{MTemp}$).} It controls the randomness introduced on top of the token predictions to mask tokens.

\paragraph{Sampling Temperature ($\mathbf{STemp}$).} It controls the randomness of the sampling from the categorical distribution of logits. Tokens are sampled from $\mathrm{logits}/\mathrm{STemp}$. STemp is calculated by \cref{eq:sample_temp}.

\begin{equation}
\mathrm{STemp} = \mathrm{bias} + \mathrm{scale} \cdot (1 - (k + 1) / K)
\label{eq:sample_temp}
\end{equation}

where $\mathrm{bias}$ and $\mathrm{scale}$ are hyperparameters (see \cref{tab:hyper}), $k$ is the current decoding iteration and $K$ is the total number of decoding iterations. We report the hyperparameters we use in in \cref{tab:hyper}.  We use bias=0.5 and scale=0.8 for all experiments.

\begin{table}[h]
\small 
\centering
\begin{center}
\resizebox{0.48\textwidth}{!}{
\begin{tabular}{l|c|ccc}
\toprule
Dataset & Method & $\mathrm{gs}$ & $\mathrm{MTemp}$ \\ %
\midrule
\multirow{2}{*}{ImageNet}
& MaskGIT++ & 65 & 6 \\ %
& \ours & 65 & 5 \\ %
& & & & \\
UCF101 & MAGVIT/ \ours& 0 & 5 \\ %
& & & & \\
\multirow{2}{*}{Kinetics600}
& MAGVIT & 10 & 12.5 \\ %
& \ours & 5 & 10 \\ %
\bottomrule
\end{tabular}
}
\caption{\small Best Sampling Hyperparameters.}
\label{tab:hyper}
\end{center}
\end{table}

\section{Additional Ablations}
\label{sec:ablation_supp}

{\flushleft \textbf{Impact of Distillation.}} We use two types of losses to train the nested sub-models - loss w.r.t the ground-truth tokens and distillation loss using the progressively bigger model as the teacher. The weight between the two losses is also linearly interpolated from the former to the latter. We compare this training strategy with the two extremes -- only ground truth loss and only distillation loss and present the results in \cref{tab:distill}. As we can see, using only distillation loss results in divergence. Using ground-truth loss is also inferior to linearly annealing on UCF101 and for the smallest model in ImageNet.

\begin{table}[h]
\small 
\centering
\begin{center}
\resizebox{0.45\textwidth}{!}{
\begin{tabular}{c| l cccc  c}
\toprule
Dataset & Training Algo. & $p=1$ & $p=2$ & $p=4$ & $p=8$ & Scheduled \\ 
\midrule
\multirow{3}{*}{ImageNet} & Only GT  & 2.5 & 3.1 & 4.1 & 6.1 & 3.1 \\
&Only Distill & \multicolumn{5}{c}{\scriptsize $\longleftarrow$ Training Diverged $\longrightarrow$ }\\
&GT $\rightarrow$ Distill & 2.5 & 3.1 & 4.1 & \textbf{5.7}  & 3.1 \\
\midrule
\multirow{3}{*}{UCF101} & Only GT  & 80.0 & 101.3 & 143.8 & 221.8 & 112.6 \\
&Only Distill & \multicolumn{5}{c}{\scriptsize $\longleftarrow$ Training Diverged $\longrightarrow$ }\\
&GT $\rightarrow$ Distill & \textbf{78.3} & \textbf{91.2} & \textbf{115.4} & \textbf{164.4} & \textbf{96.4} \\
\bottomrule
\end{tabular}
}
\end{center}
\caption{\small \textbf{Distillation Ablation.} This shows the impact of different training losses used for the nested sub-models on ImageNet256$\times$256 (size: L) and UCF101 (size: L). Using only distillation diverges while using only ground-truth losses performs worse than our approach (third row), where we combine ground-truth and distillation losses with a linear decay from the former to the latter.}
\label{tab:distill}
\end{table}

{\flushleft \textbf{Nested Attention Heads}}
We also investigate nesting along the number of attention heads ($n_{h}$), applying the same partial computation strategy as discussed before. However, this generally performed worse than nesting along the head feature dimension in attention, which is what we use for this work.

\section{Compute Gains}
\label{sec:compute_gains}

{\flushleft \textbf{Per-step FLOPs.}}
\cref{fig:flops_over_iter} illustrates the inference-time computational cost, measured in GFLOPs, per iteration for the baseline model and \ours. As we can see the amount of FLOPs can be drastically reduced using \ours. This is for a schedule with $k_1=k_2=k_3=k_4=3$. The spikes after every $3$ iterations are due to the cache refresh step. Mechanisms to get rid of the cache refresh can further reduce the total compute needed.

{\flushleft \textbf{Calculation of GFLOPs.}}
We illustrate the calculation of inference GFLOPs via Python pseudo-code in \cref{tab:flops}. We double the GFLOPs in decoding iterations where classifier-free guidance \cite{ho2022classifierfreediffusionguidance} is used. Note that we always use a cosine schedule to determine the number of tokens to be unmasked in every step.

{\flushleft \textbf{Real-Time Inference Benefits.}} In addition to the theoretical FLOP gains offered by \ours \ , here we want to analyze the real-time gains that it offers. We implement \ours \ on a single TPUv5 chip and present the results in \cref{tab:realize_gains}. 

\begin{table}[h]
\small
\centering
\resizebox{0.4\textwidth}{!}{
\begin{tabular}{l| ccc | ccc}
\toprule
Algorithm $\rightarrow$ & Baseline (MaskGIT++) & \ours \ \\ 
\midrule
Images/Sec  & 22.5 &  56.3 \\
\bottomrule
\end{tabular}
}
\caption{\small \textbf{Real-Time Inference Efficiency.}  These show the number of generated images per sec. These results are on ImageNet256$\times$256 with model size XL.}
\label{tab:realize_gains}
\end{table}

\section{Limitations.}

While our approach demonstrates strong performance in image and video generation, we acknowledge certain limitations. Some artifacts inherent to MaskGIT++ may also appear in our generated outputs (see \cref{fig:fail} for examples on ImageNet$256\times256$). Such artifacts are common in models trained on controlled datasets like ImageNet. 
Moreover, the quality of the pretrained tokenizers \citep{yu2023language, weber2024maskbitembeddingfreeimagegeneration}  directly impacts our method's effectiveness; however, improving these tokenizers is beyond the scope of this work. Although, use of nesting and decode time scaling does not have any specific requirement for model architecture and sampling scheme, KV caching requires discrete tokens.

\begin{table*}[h]
\centering
\begin{lstlisting}[language=Python, frame=single]
# Function to get the GFlops for current decoding iteration 
def get_flops(num_tokens_cached, num_tokens_processed, model_id, params, version):
    num_layers, hidden_size, mlp_dim, num_heads = params[version]
    qkv = 4 * num_tokens_processed * hidden_size * (hidden_size // model_id)
    attn = 2 * num_tokens_processed * (num_tokens_processed + num_tokens_cached) * hidden_size
    mlp = 2 * num_tokens_processed * (mlp_dim // model_id) * hidden_size
    return (qkv + attn + mlp) * num_layers // 1e9

# Function to get the total inference GFlops
def get_total_flops(version, num_iters, use_cache, refresh_cache_at, total_tokens, model_id_schedule, params, num_cond_tokens=0):
    assert num_cond_tokens < total_tokens
    refresh_cache_at = [int(x) for x in refresh_cache_at.split(',') if x]
    assert len(model_id_schedule) == num_iters
    num_cached = 0
    total_flops = 0
    
    # MaGNeTS (ours) doesn't need to process the conditioned tokens in the frame prediction task
    total_tokens -= num_cond_tokens
    
    for i in range(num_iters):
        ratio = i / num_iters
        
        # Cosine masking schedule
        num_processed = np.cos(np.pi/2. * ratio) * total_tokens

        # Even if we are performing caching, all tokens are processed in first iteration and iterations where cache is refreshed
        if i == 0 or i in refresh_cache_at and use_cache:
            total_flops += get_flops(0, total_tokens+num_cond_tokens, model_id_schedule[i], params, version)
            
        # we always cache the conditioned tokens
        else:
            total_flops += get_flops(num_cached+num_cond_tokens, total_tokens-num_cached, model_id_schedule[i], params, version)

        if use_cache:
            num_cached = total_tokens - num_processed
    return total_flops
\end{lstlisting}

\begin{lstlisting}[language=Python, frame=single]
# Sample function call for class-conditional image generation 
# params is a dictionary of the form {version: (num_layers, hidden_size, mlp_dim, num_heads)}
common = {'version': 'L', 'num_iters': 12, 'total_tokens': 257, 'params': params}
baseline = {'use_cache': False, 'refresh_at': '', 'model_id_schedule': (1,)*12, **common}
ours = {'use_cache': True, 'refresh_at': '3,6,9', 'model_id_schedule': (8,)*3+(4,)*3+(2,)*3+(1,)*3, **common}

print(get_total_flops(**baseline), get_total_flops(**ours))

# total_tokens = 1025 for class-conditional video generation and frame prediction
# num_cond_tokens = 512 for frame prediction
\end{lstlisting}
\caption{\small \textbf{Python pseudo-code illustrating the calculation of inference GFLOPs.} }
\label{tab:flops}
\end{table*}

\end{document}